\documentclass[10pt,twocolumn,letterpaper]{article}

\usepackage{cvpr}
\usepackage{times}
\usepackage{epsfig}
\usepackage{graphicx}
\usepackage{amsmath}
\usepackage{amssymb}
\usepackage{multirow}
\usepackage{booktabs}
\usepackage{comment}
\usepackage{amsmath}
\usepackage[normalem]{ulem}

\usepackage[pagebackref=true,breaklinks=true,letterpaper=true,colorlinks,bookmarks=false]{hyperref}

\cvprfinalcopy 

\def\cvprPaperID{3514} 

\ifcvprfinal\fi
\begin{document}

\title{Deep Blind Image Inpainting}

\author{Yang Liu$^{1}$, Jinshan Pan$^{2}$  , Zhixun Su$^{1}$\\
$^{1}$School of Mathematical Sciences, Dalian University of Technology\\
$^{2}$School of Computer Science and Engineering, Nanjing University of Science and Technology
}

\maketitle

\begin{abstract}
Image inpainting is a challenging problem as it needs to fill the information of the corrupted regions.
Most of the existing inpainting algorithms assume that the positions of the corrupted regions are known.
Different from the existing methods that usually make some assumptions on the corrupted regions, we present an efficient blind image inpainting algorithm to directly restore a clear image from a corrupted input.
%
%
Our algorithm is motivated by the residual learning algorithm which aims to learn the missing information in corrupted regions.
However, directly using existing residual learning algorithms in image restoration does not well solve this problem as little information is available in the corrupted regions.
To solve this problem, we introduce an encoder and decoder architecture to capture more useful information and develop a robust loss function to deal with outliers.
Our algorithm can predict the missing information in the corrupted regions, thus facilitating the clear image restoration.
Both qualitative and quantitative experimental demonstrate that our algorithm can deal with the corrupted regions of arbitrary
shapes and performs favorably against state-of-the-art methods.

\end{abstract}

\section{Introduction}

Image inpainting aims to recover a complete ideal image from the corrupted image.
The recovered region should either be as accurate as the original without disturbing uncorrupted data or visually seamlessly merged into the surrounding neighborhood such that the reconstruction result is as realistic as possible.
The applications of image inpainting are numerous, ranging from the restoration of damaged images, videos, and photographs to the removal of selected regions.
Image inpainting problem is usually modeled as
\begin{equation}
\label{eq: linear-operator}
x(i) = \left\{\begin{array}{ll} y(i) + n(i), & \mbox{}\ M(i) = 1,\\u(i),
& \mbox{}\ M(i) = 0. \end{array}\right.
\end{equation}
where $x$, $y$ and $n$ denote the corrupted image, clear image, and noise, respectively; $i$ denotes the image index; $u$ denotes the values of all other image pixels corrupted by other factors;
$M$ is a binary indicator, in which 0 denotes missing region and 1 indicates that the data is intact (which is usually influenced by noise $n$).

Recovering clear image $y$ from $x$ is a highly ill-posed problem.
Most existing inpainting algorithms \cite{Bertalmio:2000ff, Roth:2009iz, Zoran:2011jn, Ren:2015wv} usually assume that the corrupted regions are known in advance.
The success of these algorithms is mainly due to the use of different image priors~\cite{Roth:2009iz, Zoran:2011jn, Dong:2012bh}.
However, the corrupted regions are usually unknown in most cases, e.g., removing certain scratches from archived photographs.
Most existing methods cannot solve this problem well. 

Although several methods using generative adversarial nets (GANs) \cite{Goodfellow:2014wp} to solve face image inpainting problems. However, these methods cannot be applied to blind natural image inpainting as these methods highly depend on the face domain knowledge.
Thus, it is great of interest to develop an algorithm to solve the natural image inpainting when positions of corrupted regions are unknown.

%

\begin{figure}[!t]\footnotesize
	\begin{center}
		\hspace{-4.5mm}
		\begin{tabular}{ccc}
			\includegraphics*[width = 0.333\linewidth]{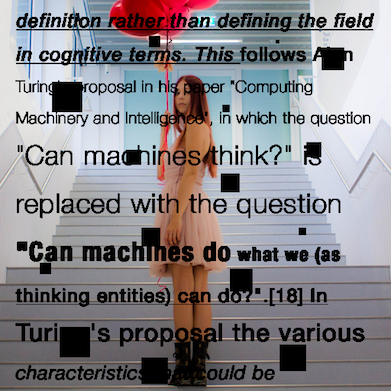} \hspace{-4.5mm}
			&\includegraphics*[width = 0.333\linewidth]{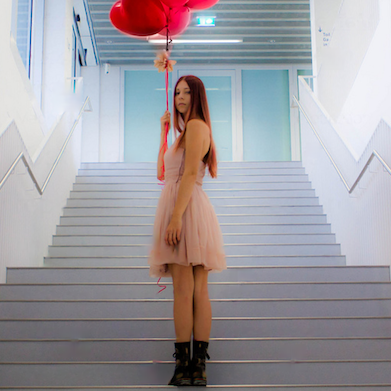} \hspace{-4.5mm}
			&\includegraphics*[width = 0.333\linewidth]{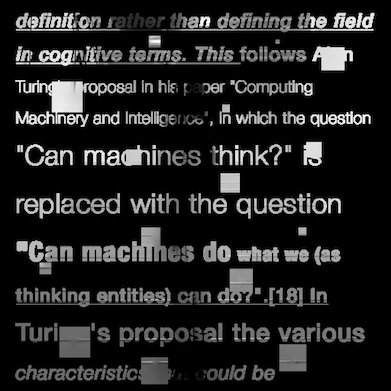} \hspace{-4.5mm}
			\\
			\scriptsize (a) Corrupted image \hspace{-4.5mm} &\scriptsize (b) Our restoration result \hspace{-4.5mm} &\scriptsize (c) Learned residual image \hspace{-4.5mm}
		\end{tabular}
	\end{center}
\vspace{-2mm}
	\caption{Image restoration using the proposed blind image inpainting method. Our method directly learns the missing information in the corrupted regions. By plugging the learned information into input image, we can get realistic images.}
	\label{figure: example for introdution}
\end{figure}

In this paper, we propose an efficient blind image inpainting algorithm to directly restore a clear image from a corrupted input using a deep convolutional neural network.
Motivated by the success of the deep residual learning algorithm~\cite{He:2016ib}, our deep feed-forward neural network learns the information that is lost in the corrupted regions.
However, as most information is missing in the corrupted regions, directly using the neural networks in image restoration, e.g.,~\cite{Kim:2016wv} does not generate clear images.
To solve this problem, we develop an encoder and decoder architecture to capture more useful information. With this architecture, our algorithm is able to predict the missing information in the corrupted regions.
With the learned information, we plug it into the input image and get the final clear image.
%
%



The main contributions of this work are summarized as follows.
\begin{itemize}
  \item We propose an effective blind image inpainting algorithm based on a deep CNN which contains an encoder and decoder architecture to capture more useful information. The proposed deep CNN is based on a residual learning algorithm which is to restore the missing information in the corrupted regions.
  \item To help the restoration of missing information, we use image gradients computed from the inputs in the residual learning.
  \item To deal with outlier, we develop a robust loss function, which ensures the training stage is stable.
  \item The proposed network is trained in an end-to-end fashion, which can handle large corrupted regions and performs favorably against state-of-the-art algorithms.
\end{itemize}

\section{Related Work}
Recent years have witnessed significant advances in image inpainting, ranging from diffusion-based algorithms and methods based on statistical priors on natural images to the striking neural network-based algorithms. We briefly review the most representative works related to ours and the applications of the convolutional neural network on image restoration.

{\flushleft {\bf Image inpainting.}} Image inpainting is first introduced in \cite{Bertalmio:2000ff}, which exploits a diffusion function to propagate low-level features from surrounding known regions to unknown regions along the boundaries of given mask. This work pioneered the idea of filling missing parts of an image using the information from surrounding regions and has been further developed by introducing the Navier-Stokes equations in another work \cite{Bertalmio:2001ev}. The normalized weighted sum of neighboring known pixels is used to infer the corrupted pixel along the image gradient in a later work \cite{Telea:2004kc}. In \cite{Bertalmio:ki}, the entire image is first decomposed into the sum of a structure part and a texture part, then these two parts are reconstructed separately based on \cite{Bertalmio:2000ff}. While these works perform well on image inpainting, global information has not been considered, and they are limited to process small objects.

Numerous image priors have been proposed to tackle image inpainting problem. In \cite{Levin:2003ge}, hand-crafted features are introduced to help to learn image statistics for inpainting. Roth and Black \cite{Roth:2009iz} develop the Fields of Experts framework based on a Markov Random Field (MRF) for learning generic image priors which encode the statistics of the entire image and could apply to general texts removal. In another work \cite{Zoran:2011jn}, patch-based Gaussian Mixture prior is proposed based on the EPLL framework which shows good performance in restoring the corrupted region. Barnes et al. \cite{Barnes:2009et} propose an efficient patch matching algorithm so the user-specified region could be filled in. These priors are powerful and present strong effectiveness in helping inpainting image, yet a problem arises, since they are designed in terms of specific regularization which itself is difficult to develop, it would be more difficult and expensive to utilize prior knowledge from various domains comprehensively.


{\flushleft {\bf CNN in image restoration.}} Recently, we have witnessed excellent progress in image restoration due to the use of CNN.
In image restoration problem, CNN can be regarded as a mapping function, which maps the corrupted images into the clear images. The main advantage of CNN is that it can be learned by large paired training data. Thus, the restoration methods using CNN achieve better performance, such as image denoising \cite{Burger:2012gm, Zhang:2017vz, Mao:2016ti}, super-resolution \cite{Dong:2016fd, Johnson:2016wm, Kim:2016wv, Shi:2016bb} and deblurring~\cite{Sun:2015we}.
One representative method is the SRCNN which is proposed by Dong et al.~\cite{Dong:2016fd}. This method proposes a three-layer CNN to reconstruct a high-resolution image from a low-resolution input.
Motivated by the success of the residual learning algorithm, Kim et al. \cite{Kim:2016wv} develop a much deeper CNN model with twenty layers based on residual learning which can generate the results with fine details.
%

Deep learning based algorithms have been made remarkable progress in image inpainting. Xie et al. \cite{Xie:2012ws} adopt a deep neural network pre-trained with sparse denoising auto-encoder and demonstrate its ability in blind image inpainting. Ren et al. \cite{Ren:2015wv} learn a three-layer convolutional neural network for non-blind image inpainting and achieve promising results given the masked image. However, this method assumes that the corrupted regions are known and it cannot be applied to the blind image inpainting problem.

Recently GANs~\cite{Goodfellow:2014wp} has been introduced into face completion. As the methods \cite{Yeh:2016wb, Li:2017vv} highly depend on the face domain knowledge, they cannot be directly extended to natural image inpainting. In addition, these approaches usually involve introducing the mask image. 

Different from existing methods, we develop a CNN based on the residual learning algorithm to solve blind image inpainting problem. Our proposed network is based on an encoder and decoder architecture, which can preserve the details and structures of the restored images.

\begin{figure*}[ht]\footnotesize
	\begin{center}	
		\begin{tabular}{cc}
			\includegraphics[width=1.0\linewidth]{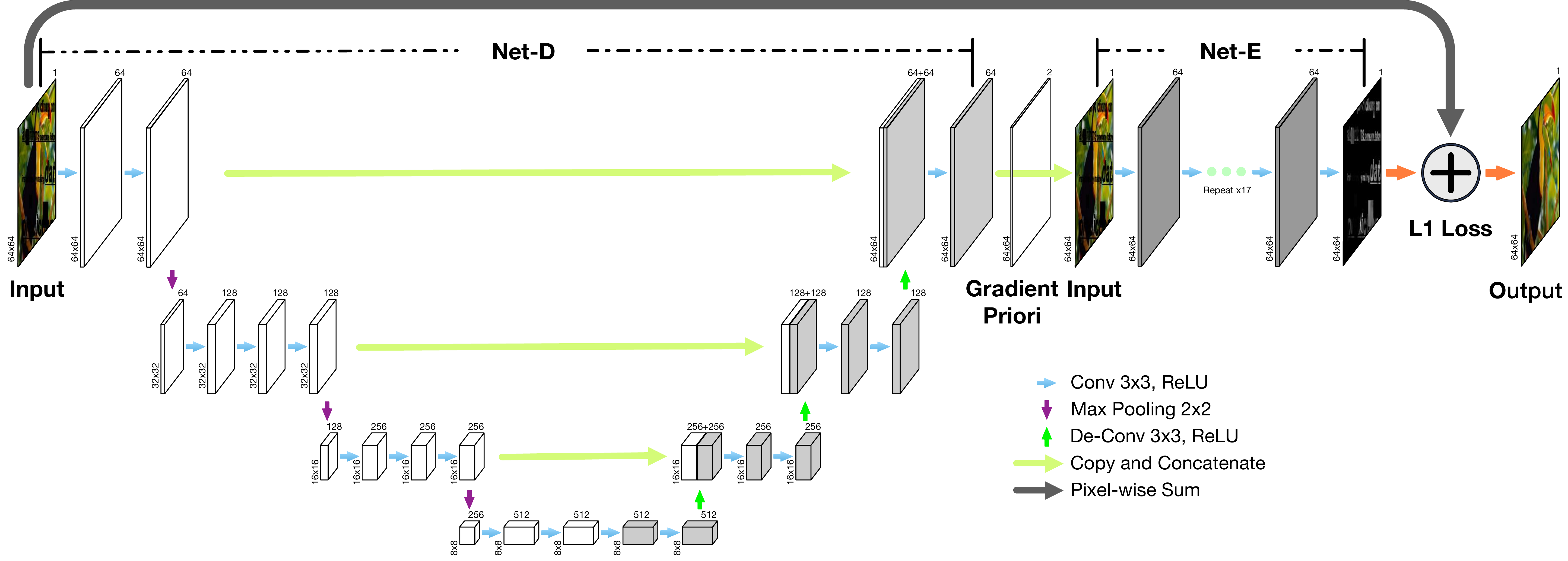}
		\end{tabular}
	\end{center}
\vspace{-2mm}
	\caption{
The proposed network architecture. Our algorithm is based on the residual learning algorithm. The net-D including encoder and decoder architecture first predicts the useful information from the input image. With the predicted features from Net-D, the Net-E takes the input image as the input image and its horizontal and vertical gradients as the input to predict the details and structures in the corrupted regions. With the predicted details and structures, we can combine the input image to estimate the final clear image.
The proposed network is jointly trained in an end-to-end manner.
}
	\label{fig: cnn framework}
\end{figure*}

\section{Proposed Algorithm}
Our algorithm is based on the residual learning algorithm. We develop a deep CNN which contains an encoder and decoder architecture to predict the missing information from the corrupted regions.
In this section, we describe the design methodology of the proposed method, including the network architecture, gradient prior, and loss functions.

%
%

\begin{table*}[h!t]\footnotesize
	\caption{Quantitative evaluations for our method and state-of-the-art methods on the benchmark datasets (Set5, Set14, Urban100, and BSDS500) in terms of PSNR and SSIM. Our method achieves significantly higher PSNR/SSIM than other methods, even non-blind inpainting methods.
	}
	\label{table: psnr/ssim - Comparisons with  State-of-the-Art Methods}
	\centering
	\vspace{2mm}
	\begin{tabular}{cccccc}
		\toprule
		\multirow{2}{*}{Dataset} &\multirow{2}{*}&FoE \cite{Roth:2009iz}&EPLL \cite{Zoran:2011jn}&ShCNN \cite{Ren:2015wv}&Ours\\
		&
		&PSNR/SSIM&PSNR/SSIM&PSNR/SSIM&PSNR/SSIM\\
		\hline
		\multirow{1}{*}{Set5}
		&\multirow{1}{*}&24.0189/0.9322&28.2710/0.9383&20.1301/0.8817&{\bfseries 32.5173/0.9569}\\
		\hline
		\multirow{1}{*}{Set14} &\multirow{1}{*}&25.5176/0.9346&28.5737/0.9393&20.9847/0.8768&{\bfseries 30.6963/0.9462}\\
		\hline
		\multirow{1}{*}{Urban100} &\multirow{1}{*}&28.1612/0.9741&30.8254/{\bfseries 0.9771}&23.9282/0.9233&{\bfseries 32.9475}/0.9748\\
		\hline
		\multirow{1}{*}{BSDS500} &\multirow{1}{*}&25.5308/0.9300&28.7912/0.9329&20.2782/0.8769&{\bfseries 30.7852/0.9440}\\
		\hline
		\bottomrule
	\end{tabular}
\end{table*}

\begin{table*}\footnotesize
	\caption{Average running time (seconds) of the evaluated methods on Set5.}
	\vspace{2mm}
	\label{tab: state-of-the-art run-time}
	\centering
	\begin{tabular}{lcccc}
		\toprule
		Methods &FoE \cite{Roth:2009iz} &EPLL \cite{Zoran:2011jn} &ShCNN \cite{Ren:2015wv} &Ours\\
		\midrule
		Avg. Running Time &336.845 &816.21 &2.12 &{\bfseries 0.03} \\
		\bottomrule
	\end{tabular}
\end{table*}

\subsection{Network Architecture}
\label{sec: network architecture}
As shown in Figure \ref{fig: cnn framework}, we design a very deep convolutional neural network concatenated by two sub-networks to learn complementary priors and further to learn the underlying mapping.
Given a single masked image to inpainting, firstly the sub-network Net-D takes it as input and outputs a multi-channel feature map which captures complementary image priors. Then these learned priors and pre-extracted gradient prior are further copied and concatenated with the masked image. After that, the sub-network Net-E takes it as input and aims to learn the underlying residual mapping \cite{He:2016ib}. Finally, the target-like restored image is yielded by pixel-wisely summing the input masked image and output residual image.

We note that the learning process of sub-network Net-D is a decoder which captures high dimension complementary image priors from low dimension input image. On the contrary, the sub-network Net-E performs the inverse transformation of Net-D which encodes the low dimension corrupted data in terms of the high dimension image information. Under these observations, we denote these two sub-networks by Net-D and Net-E.

The proposed network has 40 layers.
Net-D contains 17 convolution layers and 3 de-convolution layers, while Net-E contains 20 convolution layers. The filter size of theirs is all $3 \times 3$, each followed by a rectified linear unit(ReLU).
Inspired by the philosophy of VGG nets \cite{Simonyan:2014ws} and VDSR \cite{Kim:2016wv}, the filter number of the first layer is 64 which remains unchanged for the same output feature map size and is halved/doubled for the max pooling/de-convolution operation.
Net-D has 3 additional max pooling operations with stride 2 of size $2 \times 2$ adopted for downsampling. Besides, 3 skip link operations are used to copy and concatenate the feature maps, inspired by U-Net \cite{Ronneberger:2015vw}, which has been confirmed to favor propagating low-level context information to higher semantic layers.

We apply the zero padding of size 1 to keep all of the feature maps the same size. In this way, the residual image of the network is directly added back to the input image pixel-wisely to yield the final output image. We use this network architecture for all experiments in our work.

\subsection{Pre-extracted Gradient Prior }
We note that image gradients model the structures. Thus, we further use image gradients to help the estimation of the details and structures in the corrupted regions.

Given the corrupted image, we extract its horizontal and vertical gradients. 
These two directional gradients are then copied and concatenated with the input image to help the detail and structure estimation.
We will demonstrate the effect of the gradients in Section~\ref{ssec: gradient-prior}.


\subsection{Loss Function}
Given a training dataset $\{x^{(i)}, y^{(i)}\}_{i=1}^N$, a natural way to train the proposed network is to minimize the loss function
\begin{equation}
L(x, y) = \frac{1}{N} \sum_{i=1}^{N} \phi ((f(x^{(i)}) - y^{(i)})
\end{equation}
where $f$ is the mapping function which is learned by the proposed network, and $\phi(\cdot)$ denotes a robust function.

In this work, we take $\phi(\cdot)$ to be the $L_1$ norm instead of the $L_2$ norm to increase the robustness to the outliers.
As the network proposed is based on the residual learning algorithm \cite{He:2016ib}, the loss function used for the training is:
\begin{equation}
L(x, y) = \frac{1}{N} \sum_{i=1}^{N} \lVert(f(x^{(i)}) + x^{(i)} - y^{(i)}\rVert_1
\label{formula: L1 loss function}
\end{equation}
We will show that the loss function based on $L_1$ norm is more robust to outliers thereby leading to higher accuracy and visual quality in Section~\ref{ssec: Effect of the Loss Function}.

\section{Experimental Results}
In this section, we evaluate our algorithm on several benchmark datasets and compare it with state-of-the-art methods. 
Similar to existing methods~\cite{Xie:2012ws, Ren:2015wv}, we only train our model on grayscale images, and all the PSNR and SSIM values in this paper are computed by grayscale images. 
We use color images for visualization, where each channel of the color image is processed independently. 
The trained models and source code will be made available to the public.

\begin{figure*}[ht]\footnotesize
	\begin{center}
		\begin{tabular}{cccccc}
			\includegraphics[width = 0.16\linewidth]{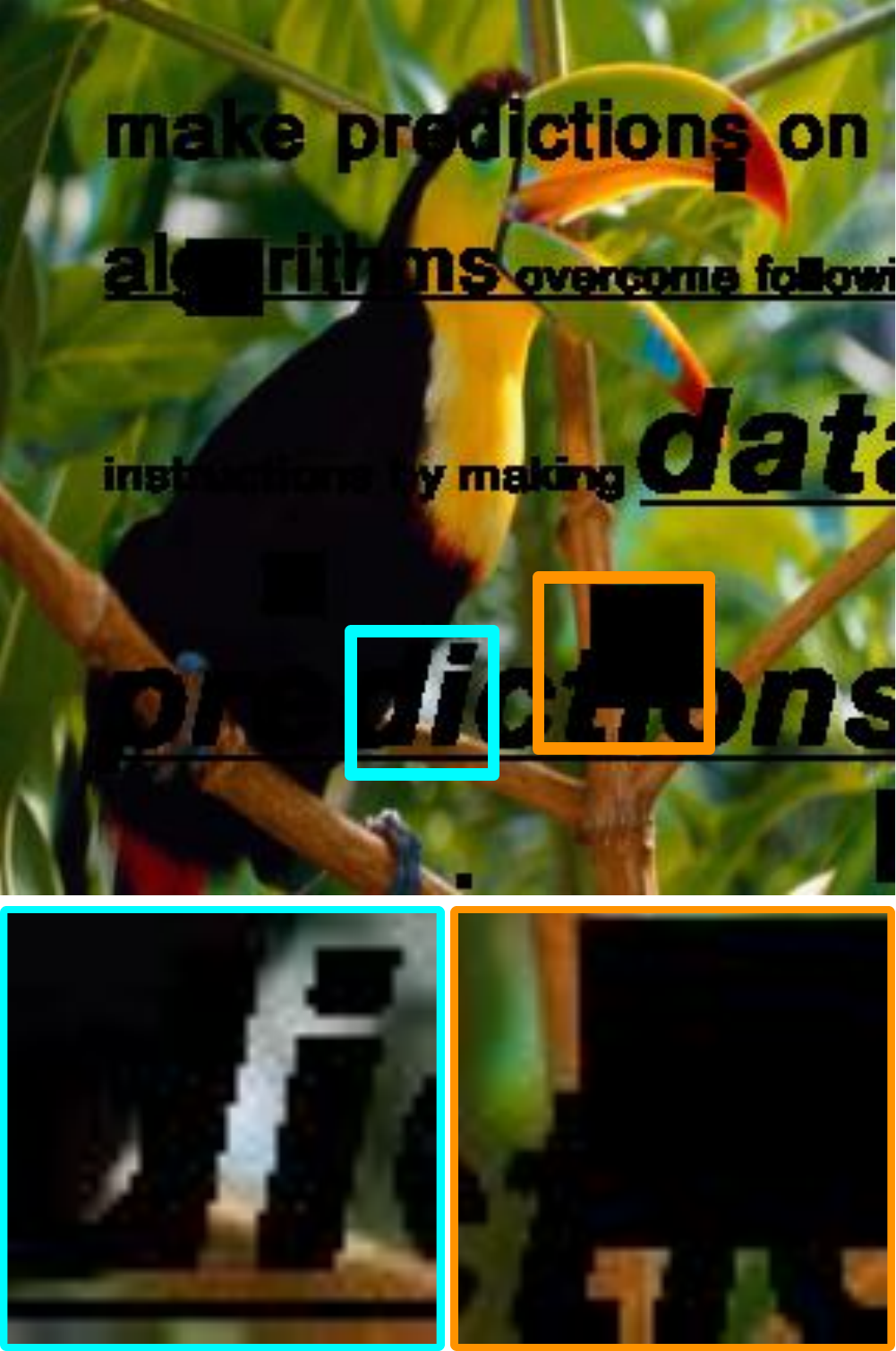} \hspace{-4mm}
			&\includegraphics[width = 0.16\linewidth]{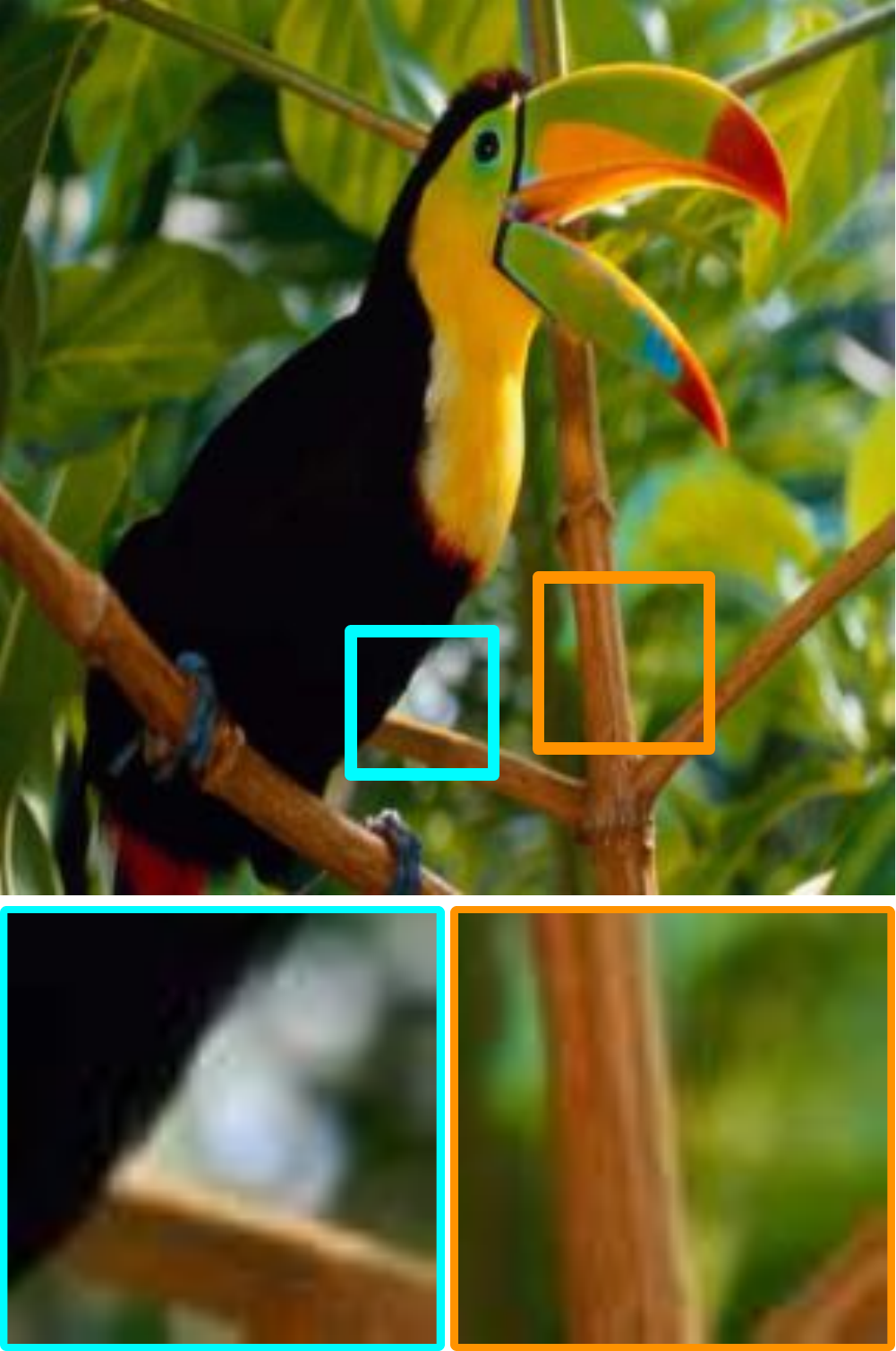} \hspace{-4mm}
			&\includegraphics[width = 0.16\linewidth]{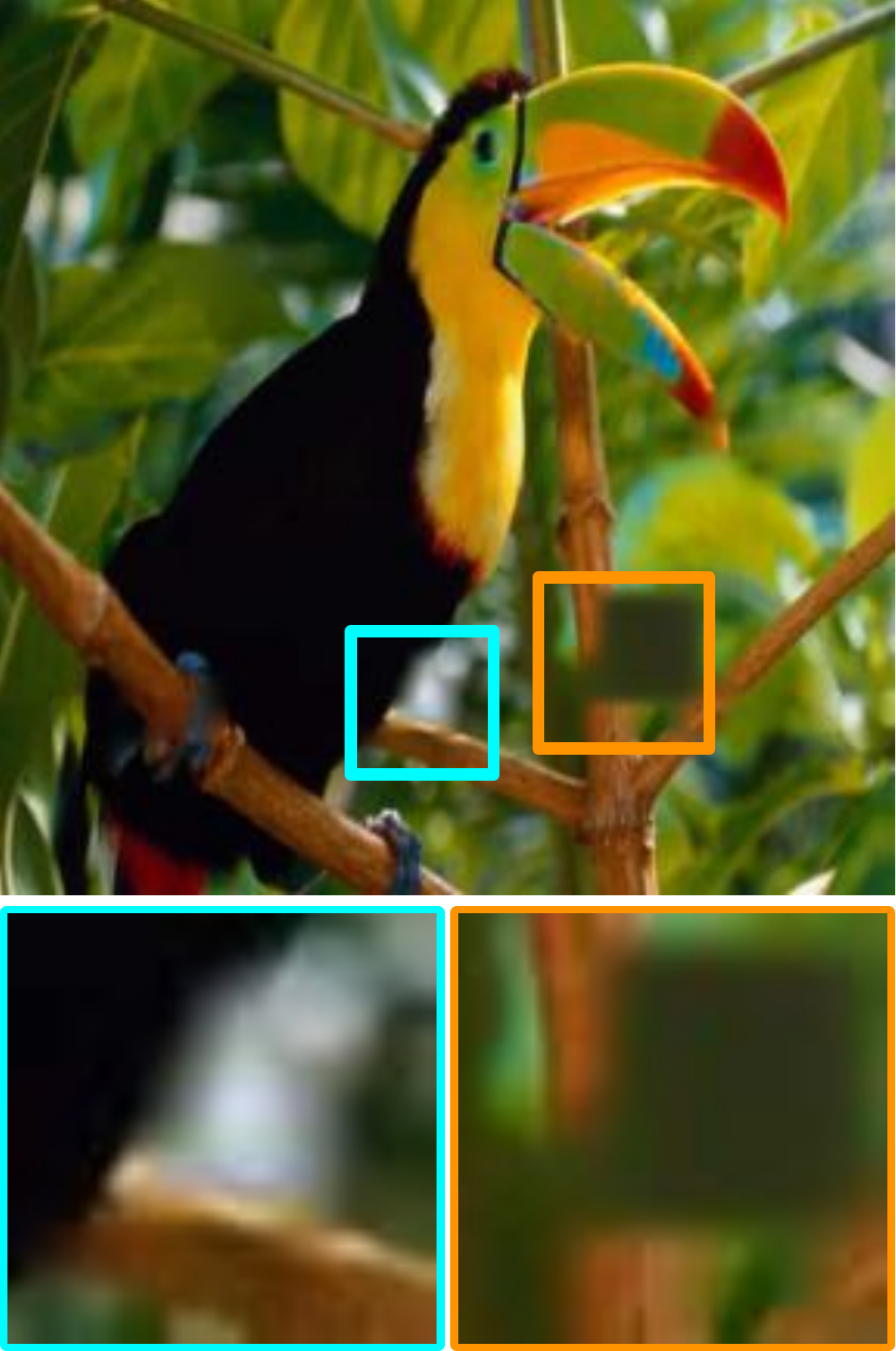} \hspace{-4mm}
			&\includegraphics[width = 0.16\linewidth]{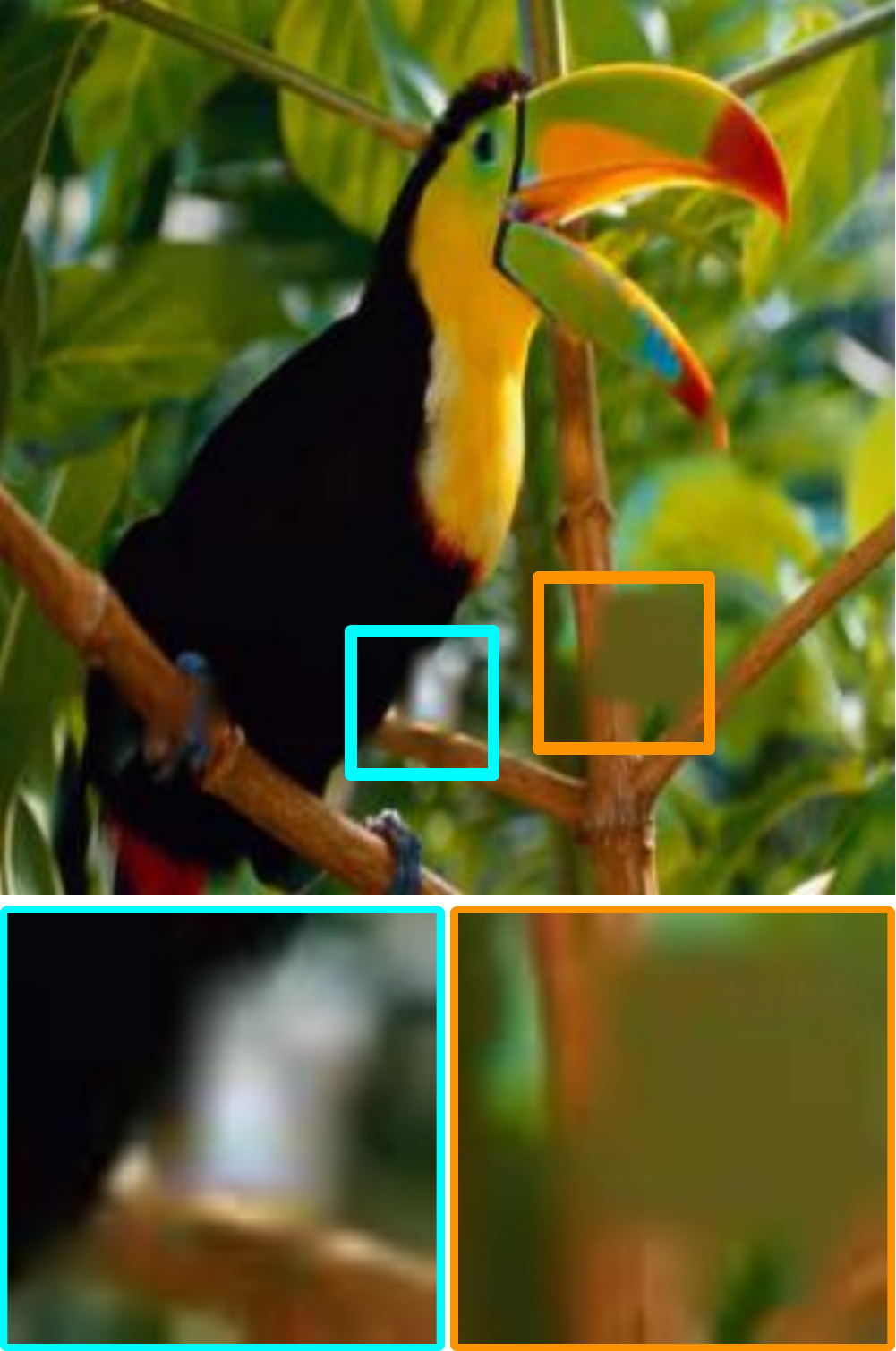} \hspace{-4mm}
			&\includegraphics[width = 0.16\linewidth]{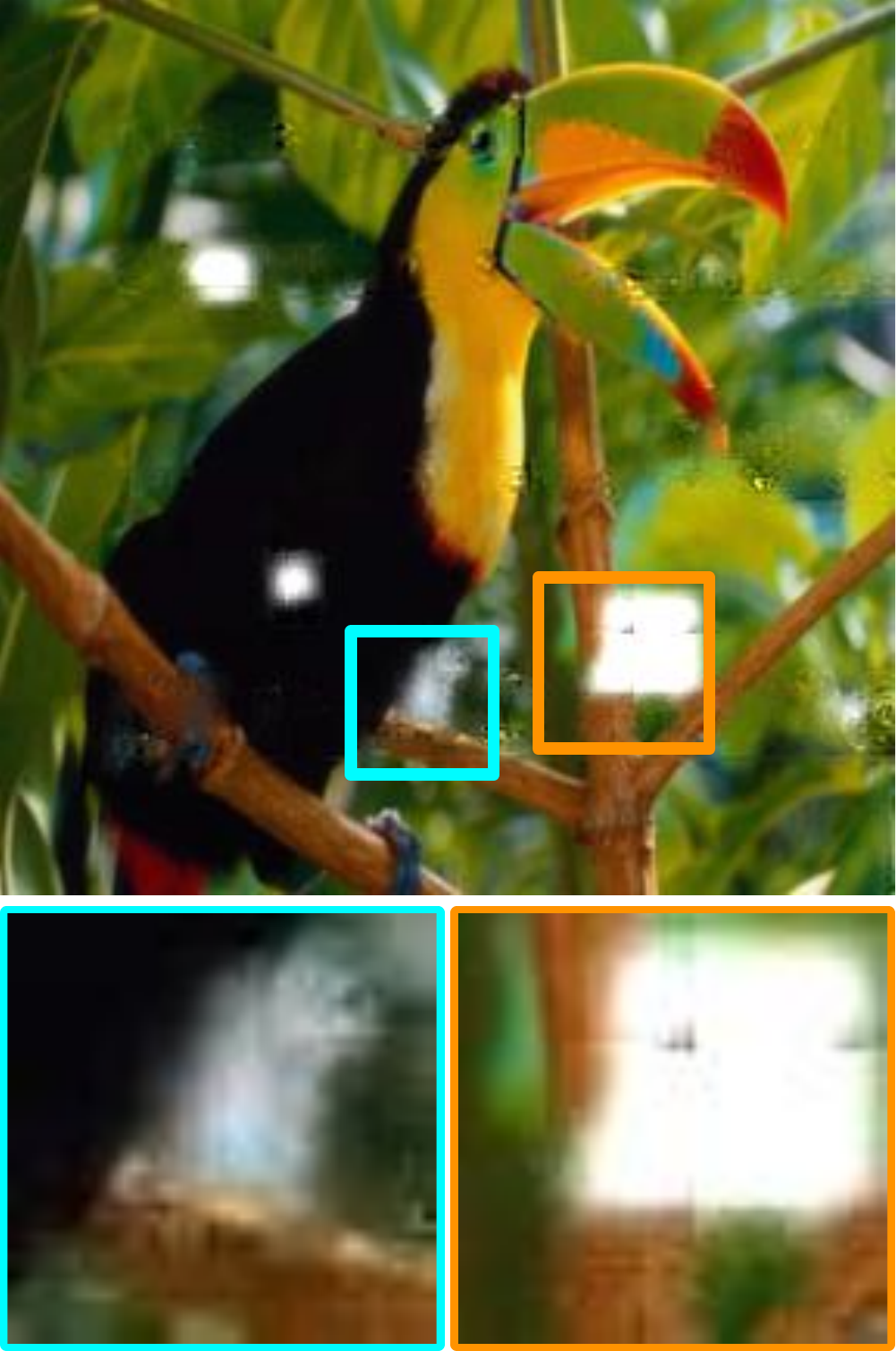} \hspace{-4mm}
			&\includegraphics[width = 0.16\linewidth]{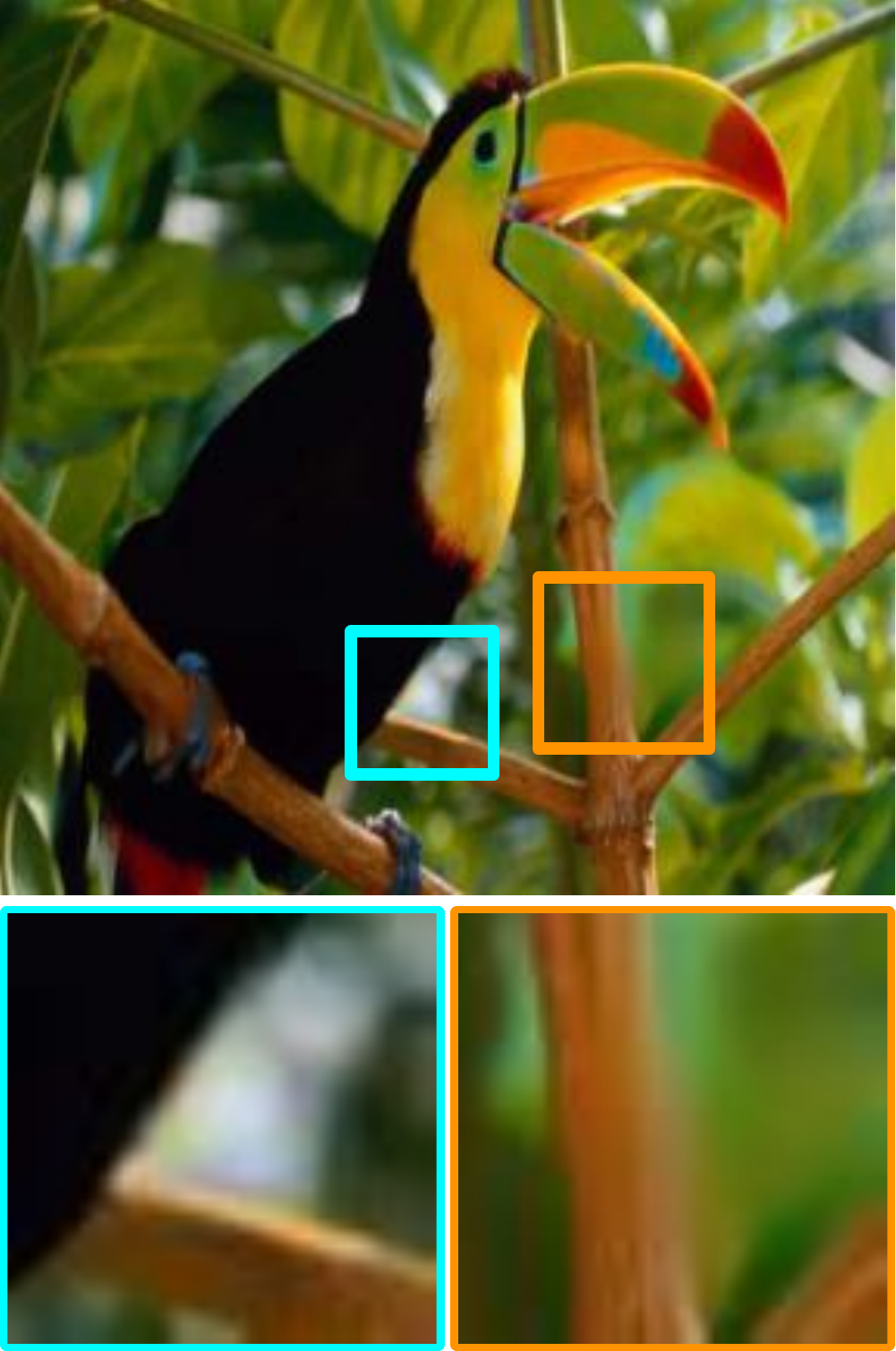} \hspace{-4mm}
			\\
			(PSNR, SSIM) \hspace{-4mm} &(-, -) \hspace{-4mm} &(30.4127, 0.9661) \hspace{-4mm} &(32.5860, 0.9650) \hspace{-4mm} &(20.9166, 0.9030) \hspace{-4mm} &({\bfseries 35.9517}, {\bfseries 0.9784}) \hspace{-4mm}
			\\
            \includegraphics[width = 0.16\linewidth]{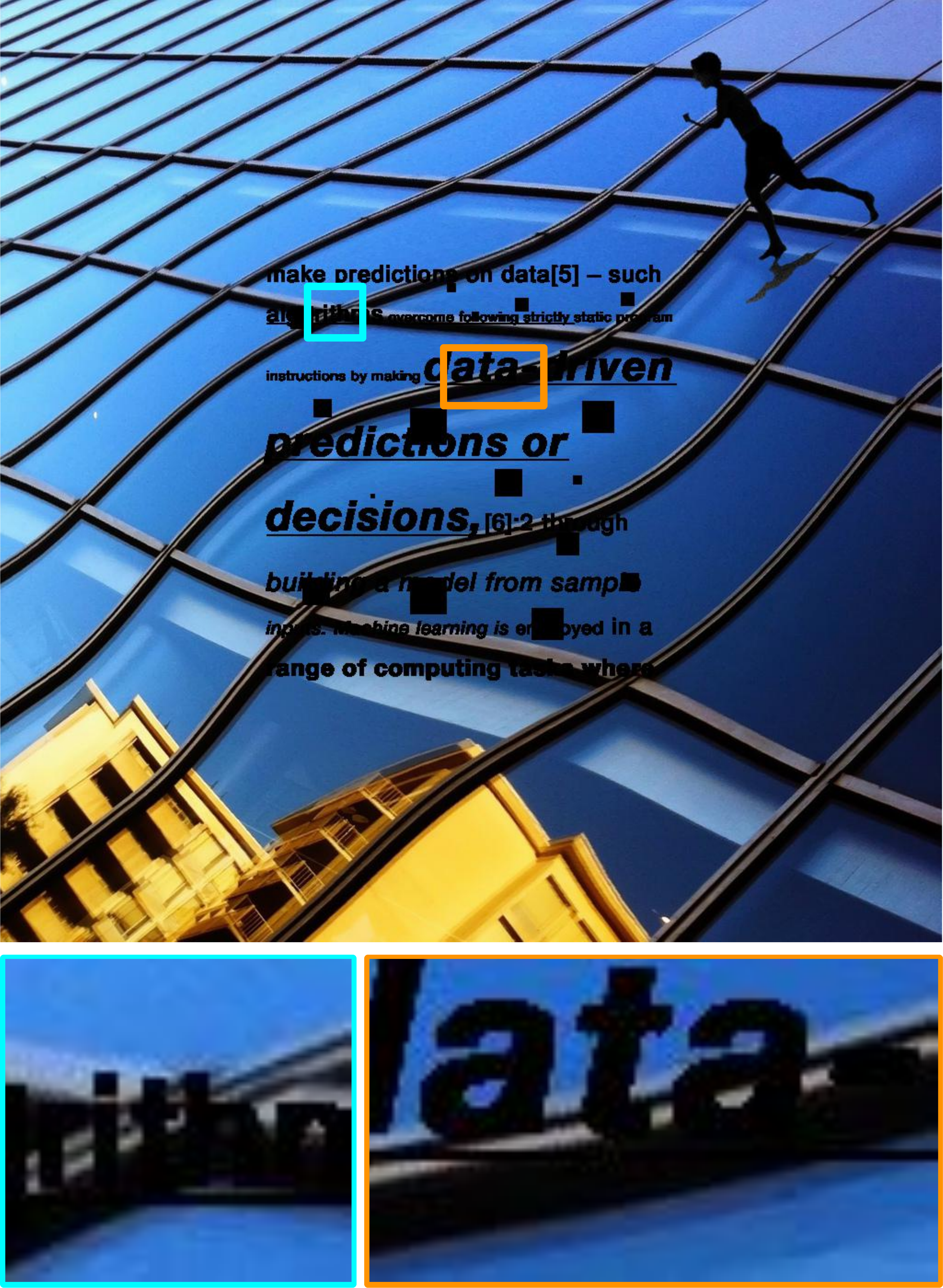} \hspace{-4mm}
			&\includegraphics[width = 0.16\linewidth]{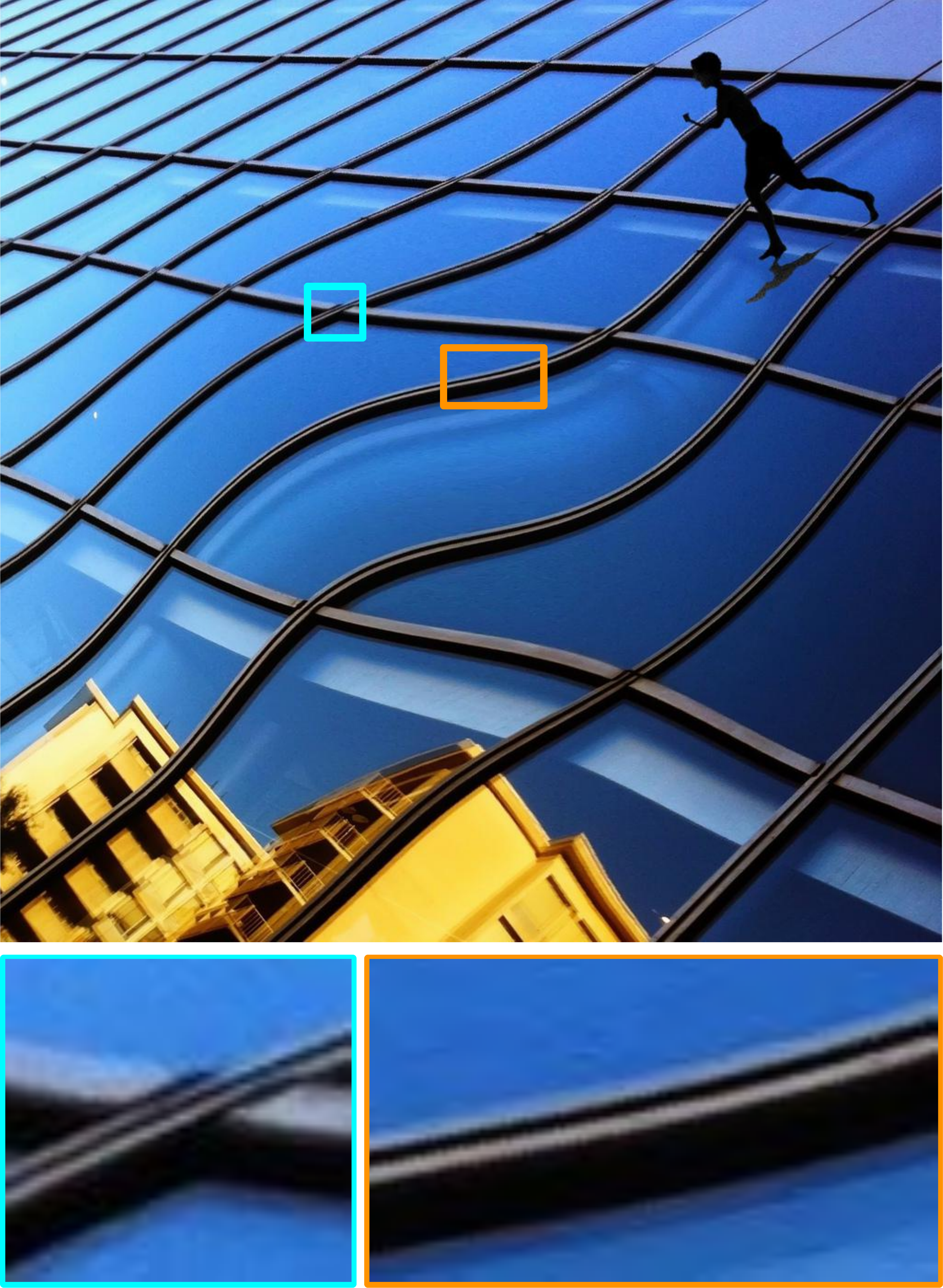} \hspace{-4mm}
			&\includegraphics[width = 0.16\linewidth]{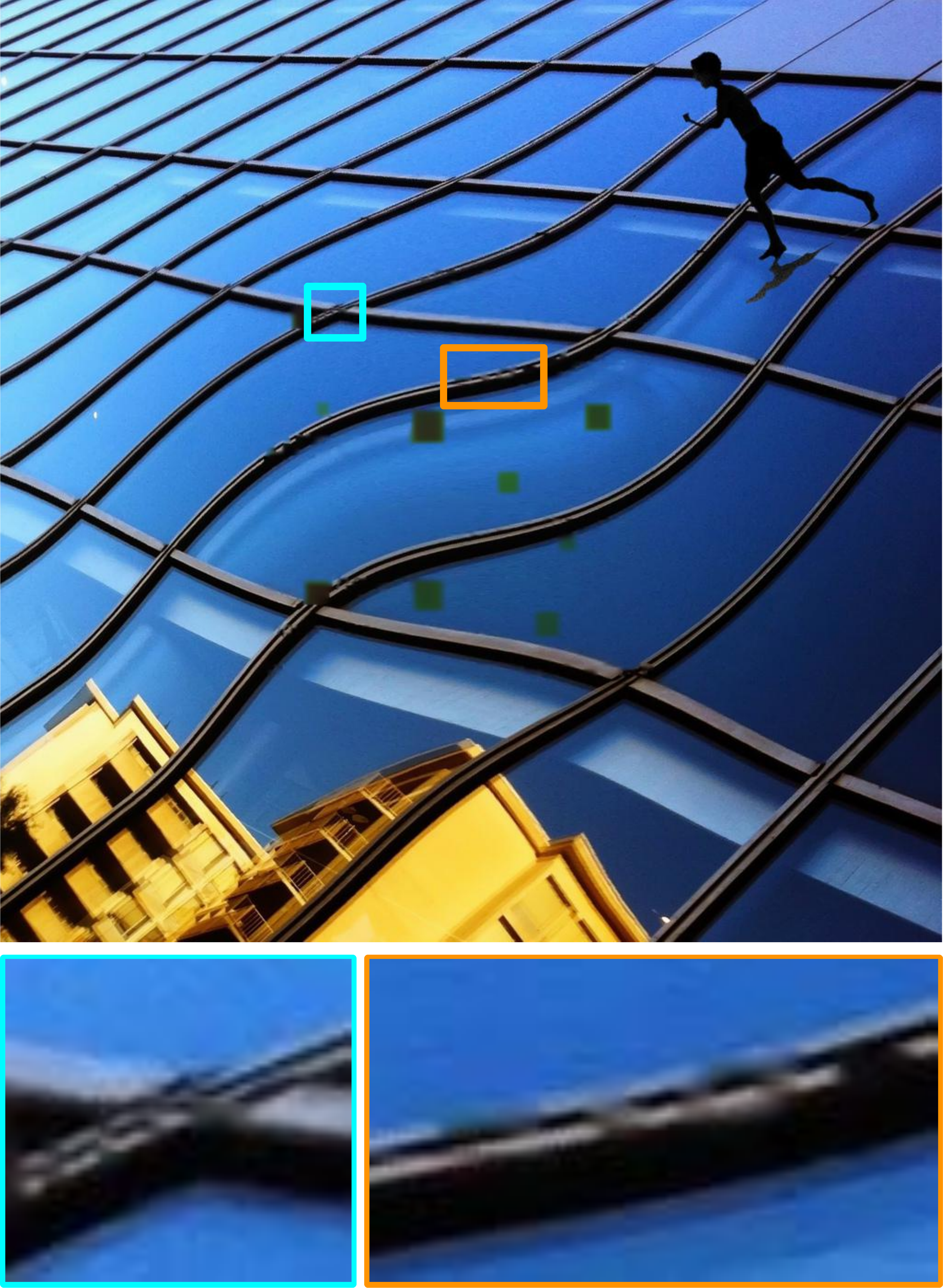} \hspace{-4mm}
			&\includegraphics[width = 0.16\linewidth]{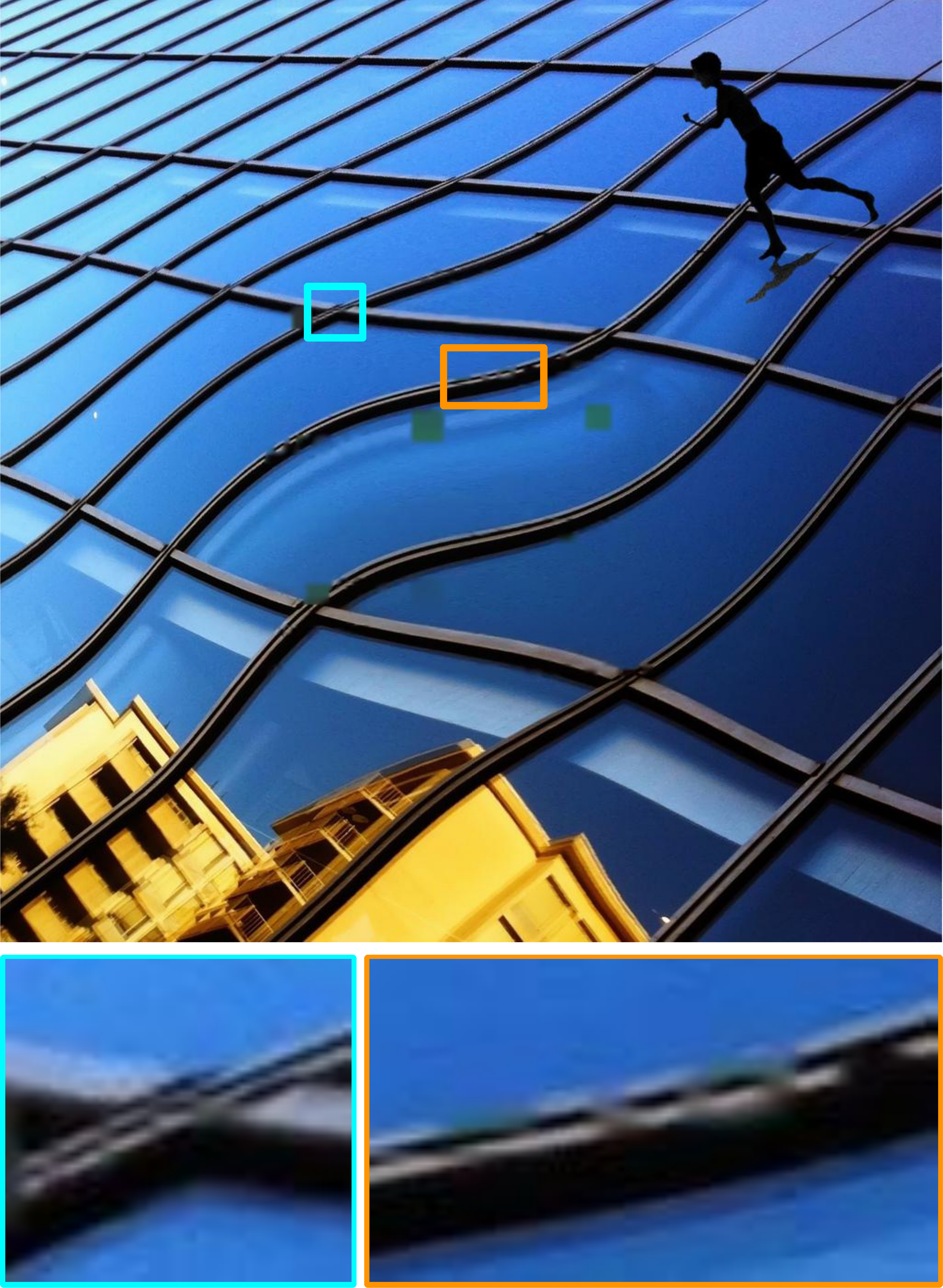} \hspace{-4mm}
			&\includegraphics[width = 0.16\linewidth]{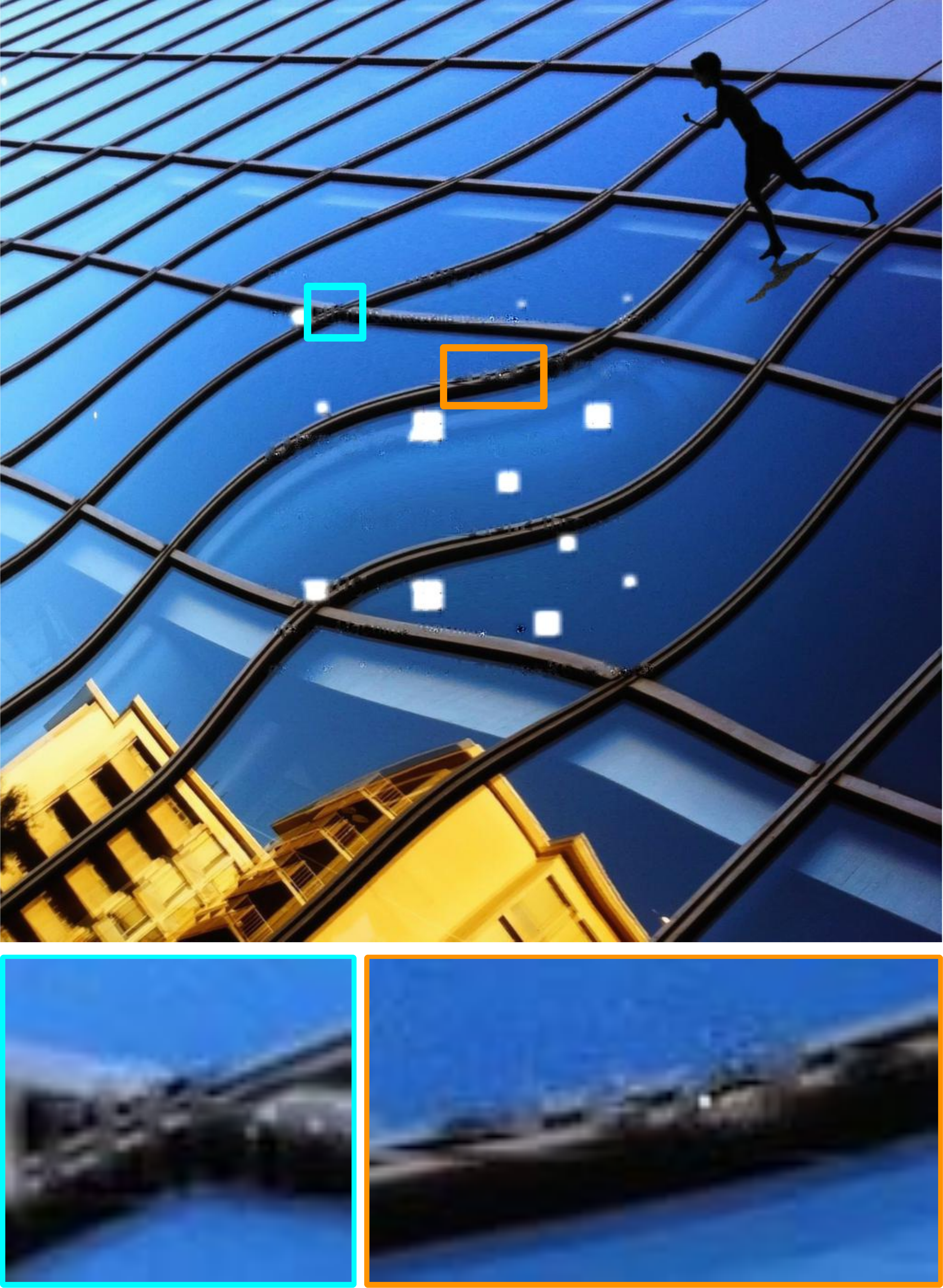} \hspace{-4mm}
			&\includegraphics[width = 0.16\linewidth]{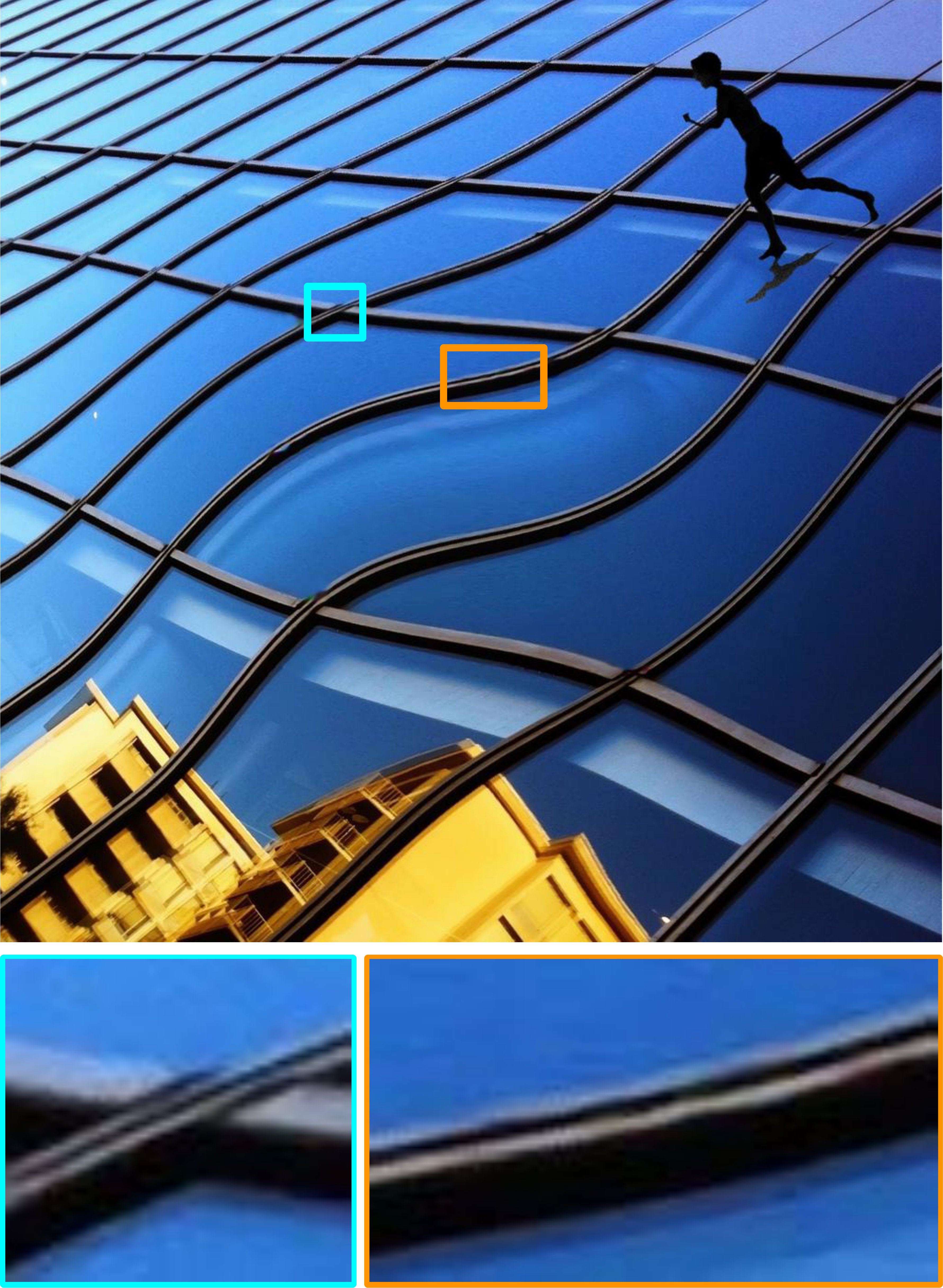} 
			\hspace{-4mm}
			\\
			(PSNR, SSIM) \hspace{-4mm} &(-, -) \hspace{-4mm} &(37.3424, 0.9938) \hspace{-4mm} 
			&(39.4181, 0.9940) \hspace{-4mm} &(25.7797, 0.9587) \hspace{-4mm} &({\bfseries 43.4470}, {\bfseries 0.9963}) \hspace{-4mm}
			\\
			\includegraphics[width = 0.16\linewidth]{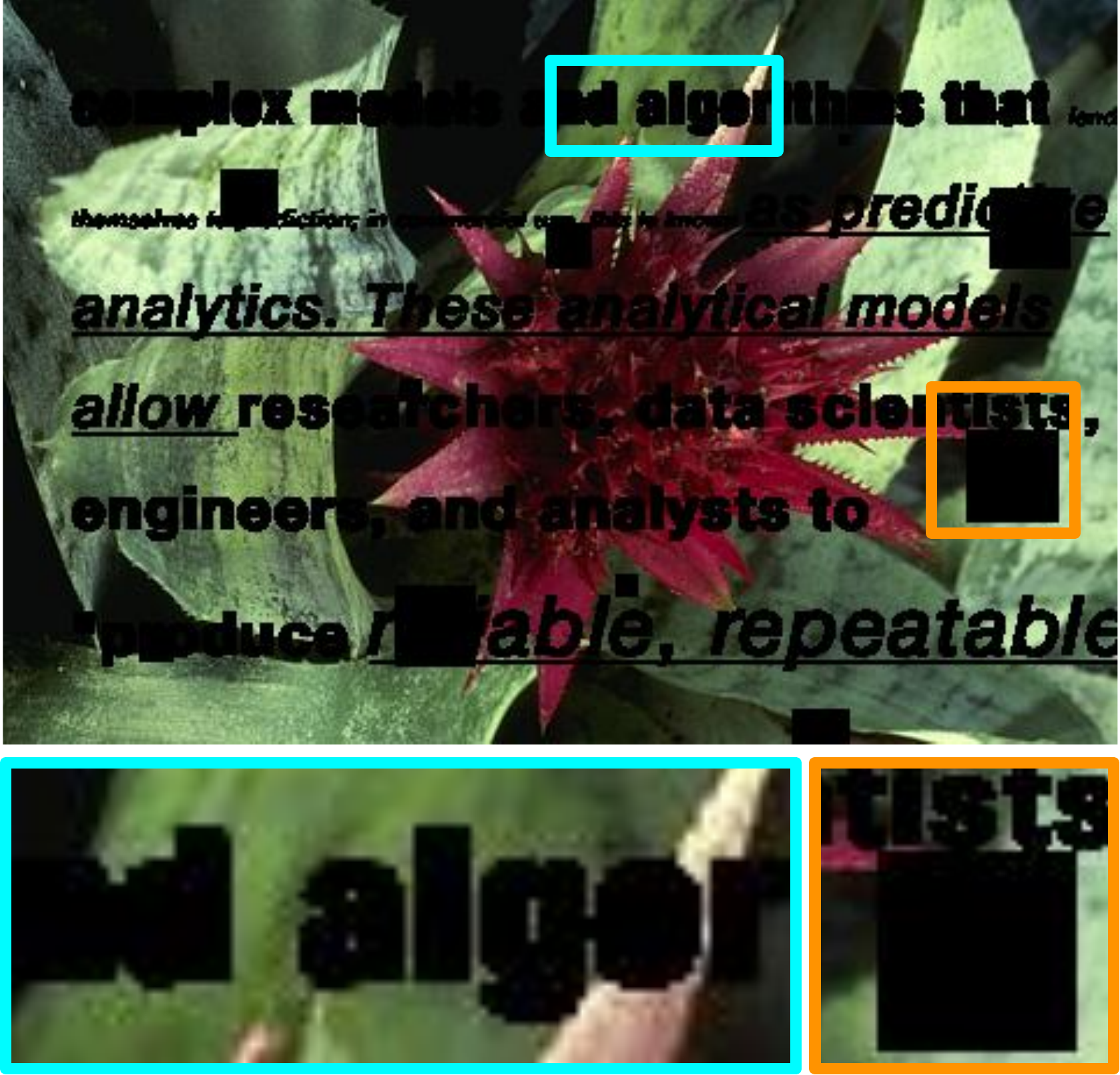} \hspace{-4mm}
			&\includegraphics[width = 0.16\linewidth]{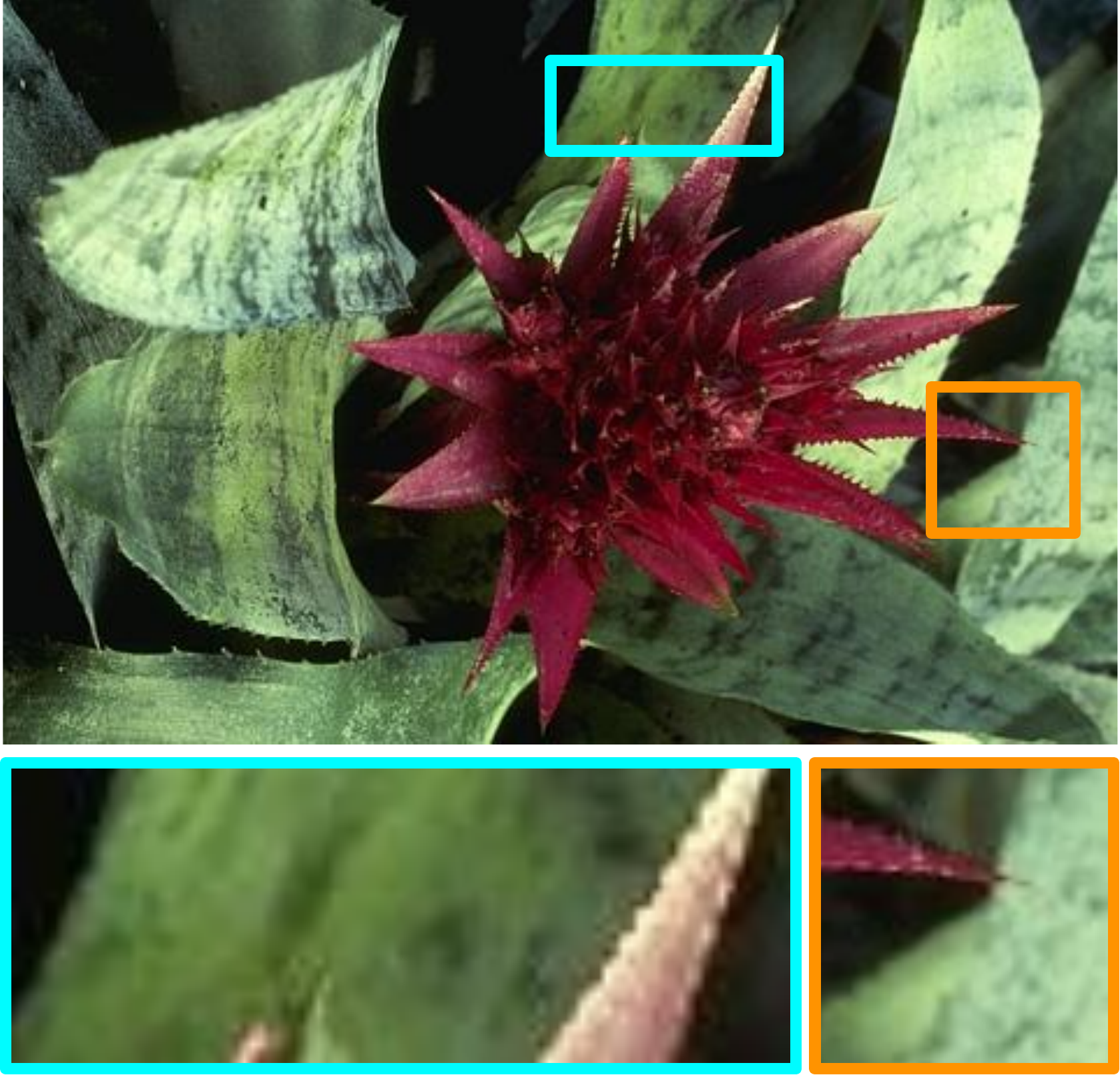} \hspace{-4mm}
			&\includegraphics[width = 0.16\linewidth]{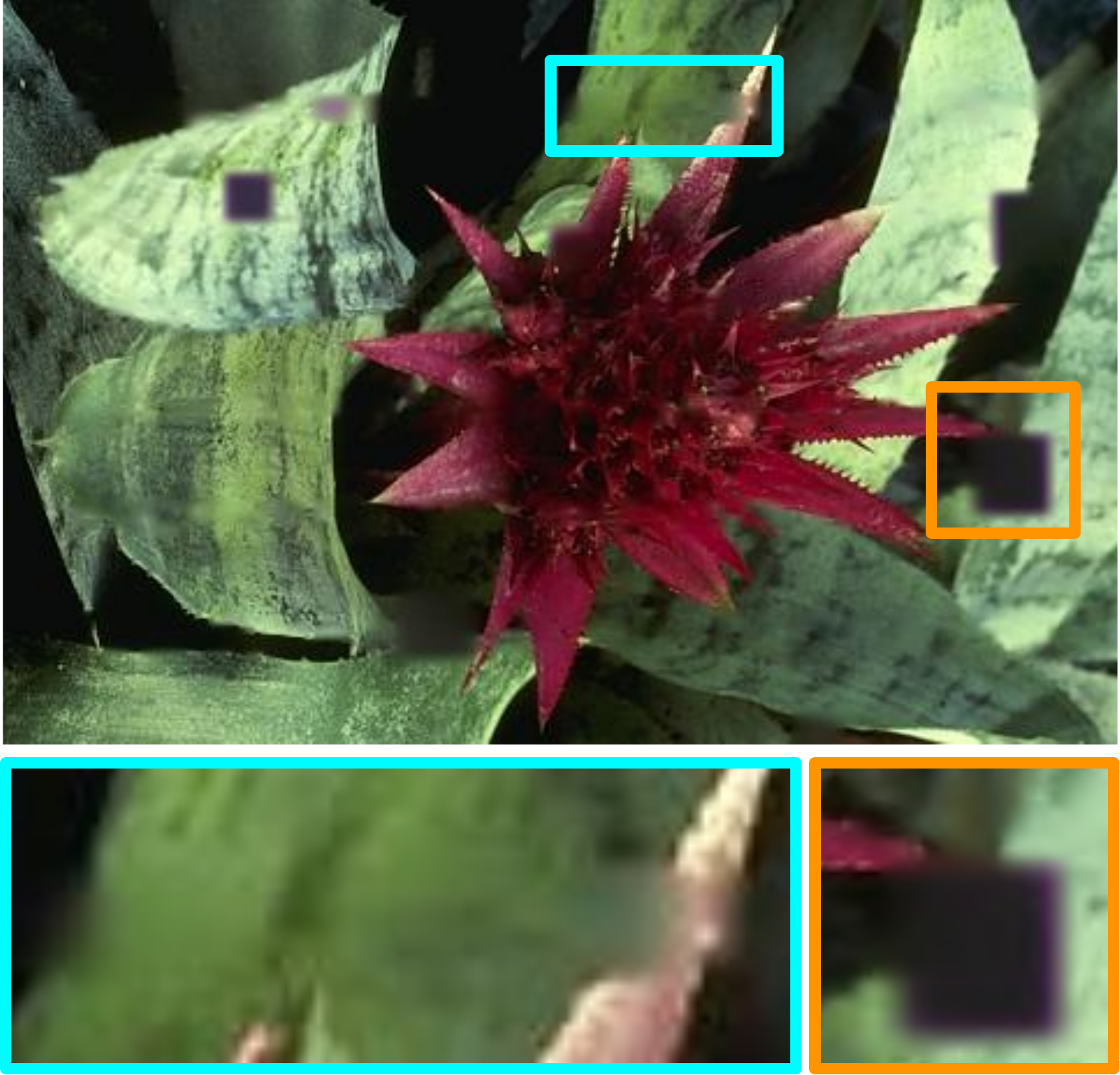} \hspace{-4mm}
			&\includegraphics[width = 0.16\linewidth]{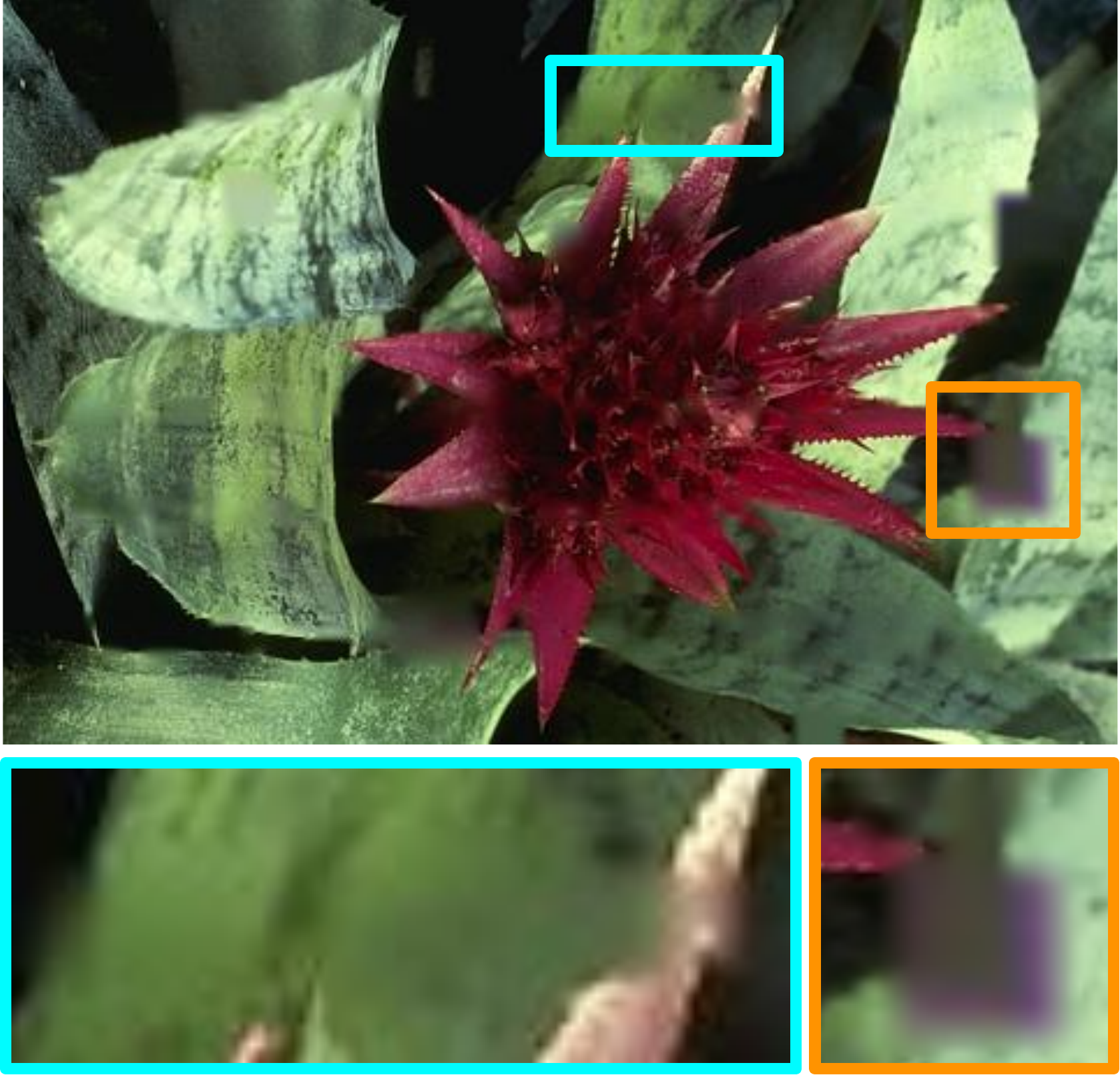} \hspace{-4mm}
			&\includegraphics[width = 0.16\linewidth]{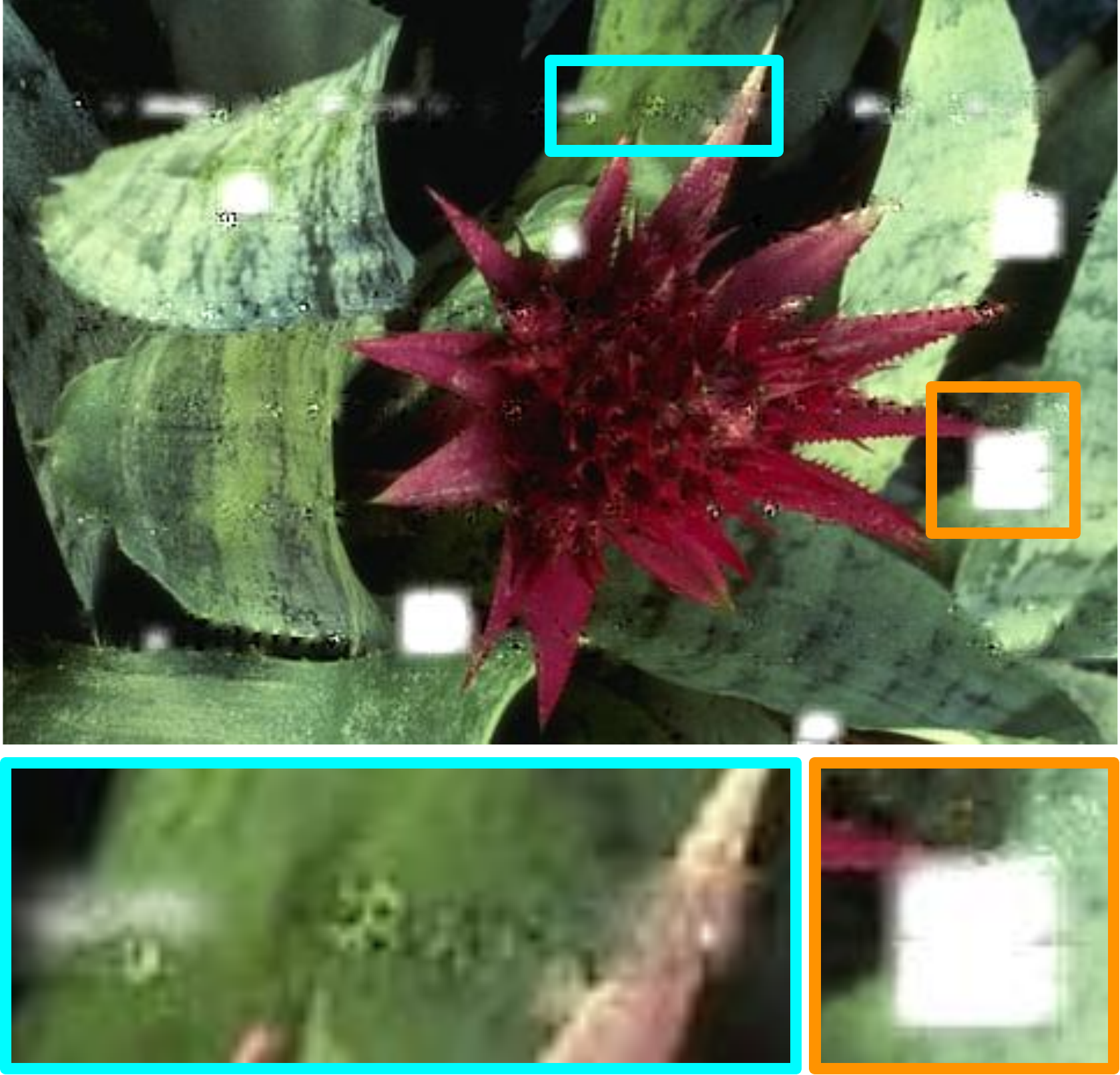} \hspace{-4mm}
			&\includegraphics[width = 0.16\linewidth]{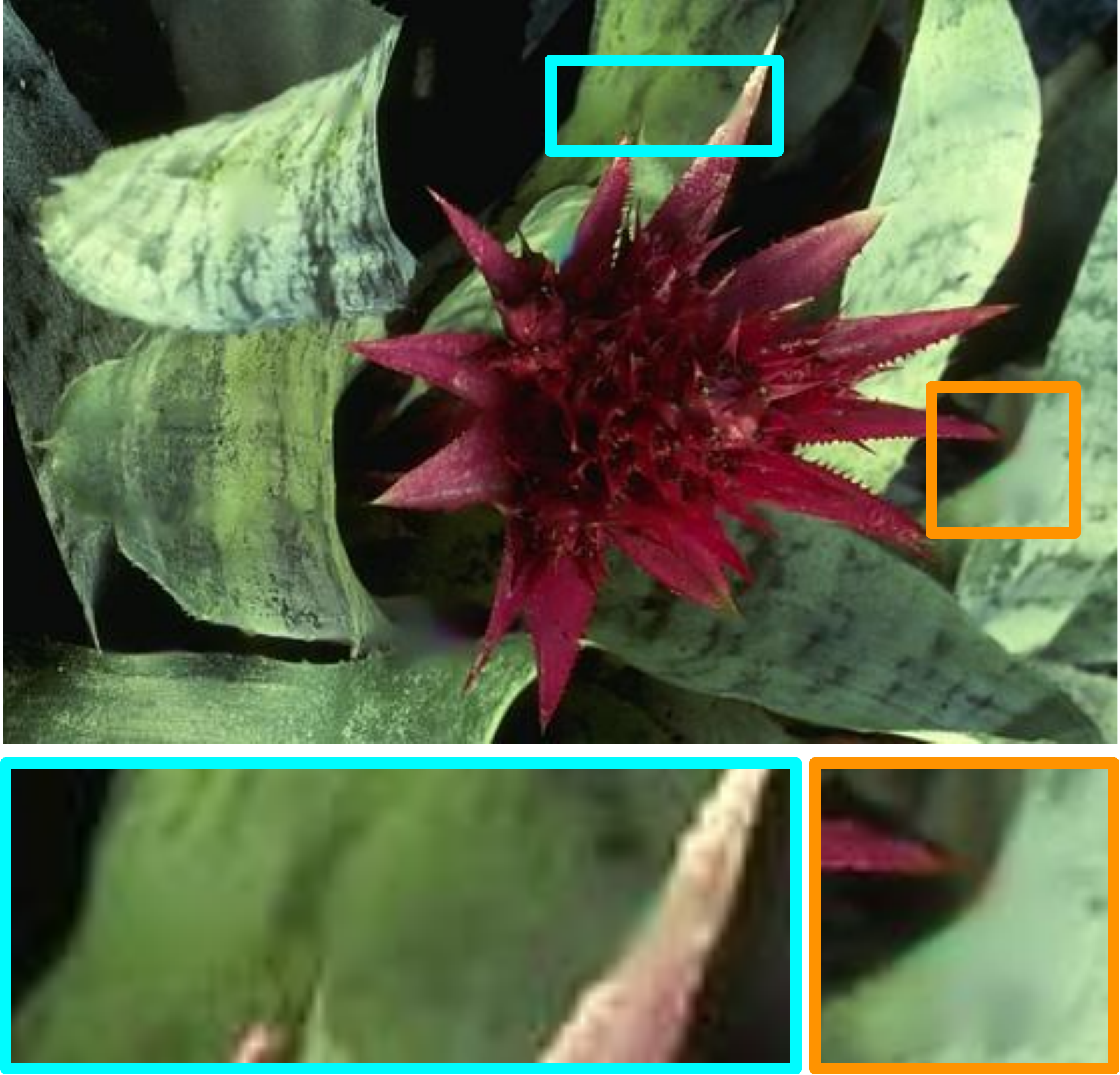} 
			\hspace{-4mm}
			\\
			(PSNR, SSIM) \hspace{-4mm} &(-, -) \hspace{-4mm} &(23.7456, 0.9186) \hspace{-4mm} 
			&(25.9831, 0.9157) \hspace{-4mm} &(19.3231, 0.8457) \hspace{-4mm} &({\bfseries 30.9726}, {\bfseries 0.9373}) \hspace{-4mm}
			\\
			(a) Input \hspace{-4mm} &(b) Ground Truth \hspace{-4mm} &(c) FoE \cite{Roth:2009iz} \hspace{-4mm} &(d) EPLL \cite{Zoran:2011jn} \hspace{-4mm} &(e) ShCNN \cite{Ren:2015wv} \hspace{-4mm} &(f) Ours \hspace{-4mm}
			\\
		\end{tabular}
%
%
	\end{center}
	\caption{Visual comparisons with state-of-the-art methods.
		Previous blind and non-blind inpainting methods fail to recover missing data of square regions and contain considerable artifacts in dealing with texts, while the proposed method achieves considerably pleasing visual quality and much higher accuracy.
	}
	\label{fig: state-of-the-art visual comparisons}
\end{figure*}

\begin{figure*}[ht]\footnotesize
	\begin{center}
		\begin{tabular}{cccccc}
			\includegraphics[width = 0.16\linewidth]{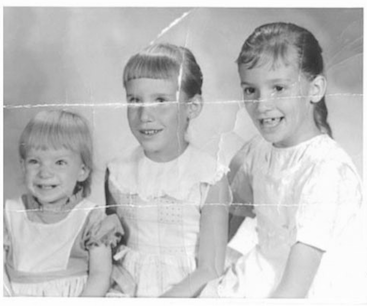} \hspace{-4.5mm}
			&\includegraphics[width = 0.16\linewidth]{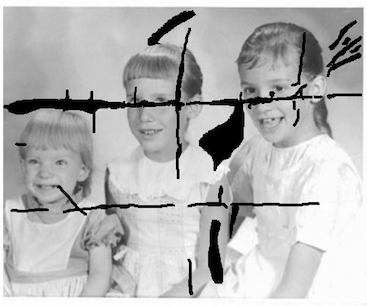} \hspace{-4.5mm}
			&\includegraphics[width = 0.16\linewidth]{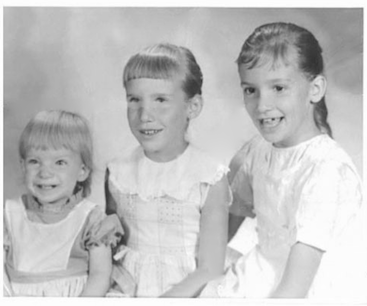} \hspace{-4.5mm}
			&\includegraphics[width = 0.16\linewidth]{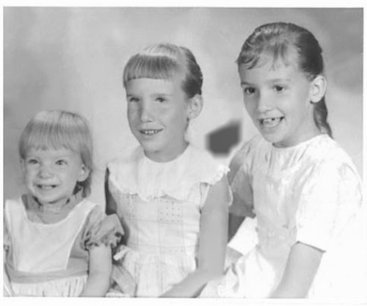} \hspace{-4.5mm}
			&\includegraphics[width = 0.16\linewidth]{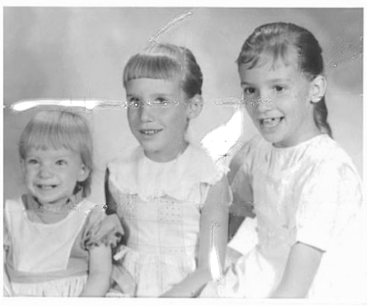} \hspace{-4.5mm}
			&\includegraphics[width = 0.16\linewidth]{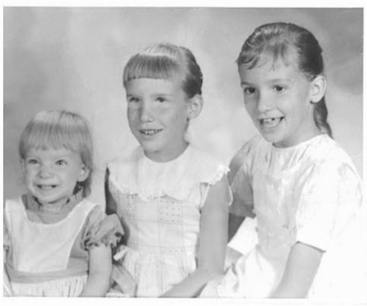} \hspace{-4.5mm}
			\\
			(a) Input \hspace{-4.5mm} &(b) Masked image  \hspace{-4.5mm} &(c) FoE \cite{Roth:2009iz} \hspace{-4.5mm} &(d) EPLL \cite{Zoran:2011jn} \hspace{-4.5mm} &(e) ShCNN \cite{Ren:2015wv} \hspace{-4.5mm} &(f) Ours \hspace{-4.5mm}
			\\
		\end{tabular}
	\end{center}
\caption{Visual comparisons with state-of-the-art methods on an old photograph. Our algorithm does not require the positions of corrupted regions and generates comparable results.
}
\label{fig: state-of-the-art visual comparison on an old photograph}
\end{figure*}

\subsection{Datasets}
We create a binary mask image dataset which contains 20 text images for generating training dataset and another 10 text images for generating test dataset with image size of $512 \times 512$. Randomly distributed squares with the size of 5 to 40 pixels are added.
The texts are of the most popular font Arial and Helvetica Neue with different font sizes of 10 to 50 pt and the most common font styles, including bold, thin, italic, light, ultralight, underline, condensed styles and their combinations. Note that the texts and squares indicate the corrupted regions which are unknown in advance in blind inpainting task.

{\flushleft \textbf{Training dataset.}} We generate the ground truth of the training data by randomly sampled 100,000 $64 \times 64$ grayscale patches from 291 natural images \cite{Martin:jm} with data augmentation, e.g., scale and rotation. Then the masked patches are generated by taking a pixel-wise multiplication between the ground truth and the created 20 text images for training, plus randomly distributed squares.

{\flushleft \textbf{Test dataset.}} Four datasets are employed for evaluating the proposed algorithm, i.e., Set5, Set14, Urban100 \cite{Huang:2015hy} and BSDS500 \cite{Arbelaez:2011jg}, which have been widely used as benchmark test datasets in image restoration problems. The masked test images are generated by taking the same pixel-wise multiplication with the created 10 text images for test, plus randomly distributed squares which are identical for each test method.

\subsection{Network Training}
\label{section: Training Details and Parameters}
Our implementation is based on PyTorch. We trained all networks on an NVIDIA 1080 Ti GPU using the created 100,000 image patches of the size $64 \times 64$ pixels.
The weights of our proposed network are initialized by the method proposed by He et al. \cite{He:2015dj}, which is confirmed to favor the performance of the rectified linear unit (ReLU).
We take Adam \cite{Kingma:2014us} as the optimization method with $\beta = (0.9, 0.999)$ and set the weight decay to be 0. The number of the optimal training epoch is 60 (47100 iterations in total with batch size 128). The network is first trained with 50 epochs at a learning rate of $10^{-3}$, then another 10 epochs at a lower rate of $10^{-4}$. Training with too many iterations has no improvement for the results in terms of our observations.

\begin{table*}[tp]\footnotesize
	\caption{Quantitative evaluations for the different part of our network on the benchmark datasets (Set5, Set14, Urban100, and BSDS500) in terms of PSNR and SSIM, where GP is short for gradient prior. The complete network (Net-D+GP+Net-E+$L_1$) achieves significantly higher PSNR/SSIM than other cases.
	}
	\label{table: psnr/ssim of different networks}
	\centering
	\vspace{2mm}
	\begin{tabular}{ccccccc}
		\toprule
		\multirow{2}{*}{Dataset} &\multirow{2}{*}& Net-D+GP+Net-E+$L_1$& GP+Net-E+$L_1$& Net-E$^{64}$+GP+Net-E+$L_1$& Net-D+Net-E+$L_1$& Net-D+GP+Net-E+$L_2$\\
		&
		&PSNR/SSIM&PSNR/SSIM&PSNR/SSIM&PSNR/SSIM&PSNR/SSIM\\
		\hline
		\multirow{1}{*}{Set5}
		&\multirow{1}{*}&{\bfseries 32.5173/0.9569}&31.9124/0.9532&31.5729/0.9525&32.0818/0.9532&31.9615/0.9509\\
		\hline
		\multirow{1}{*}{Set14} &\multirow{1}{*}&{\bfseries 30.6963/0.9462}&30.0898/0.9434&30.1323/0.9437&30.5271/0.9441&30.1705/0.9410\\
		\hline
		\multirow{1}{*}{Urban100} &\multirow{1}{*}&{\bfseries 32.9475/0.9748}&32.4121/0.9729&32.3807/0.9732&32.8506/0.9728&32.6586/0.9704\\
		\hline
		\multirow{1}{*}{BSDS500} &\multirow{1}{*}&{\bfseries 30.7852/0.9440}&30.5307/0.9418&30.5135/0.9415&30.6858/0.9424&30.6189/0.9407\\
		\hline
		\bottomrule
	\end{tabular}
\end{table*}

\begin{figure*}[htp]\footnotesize
	\begin{center}
		\begin{tabular}{ccccc}
			\includegraphics[width = 0.19\linewidth]{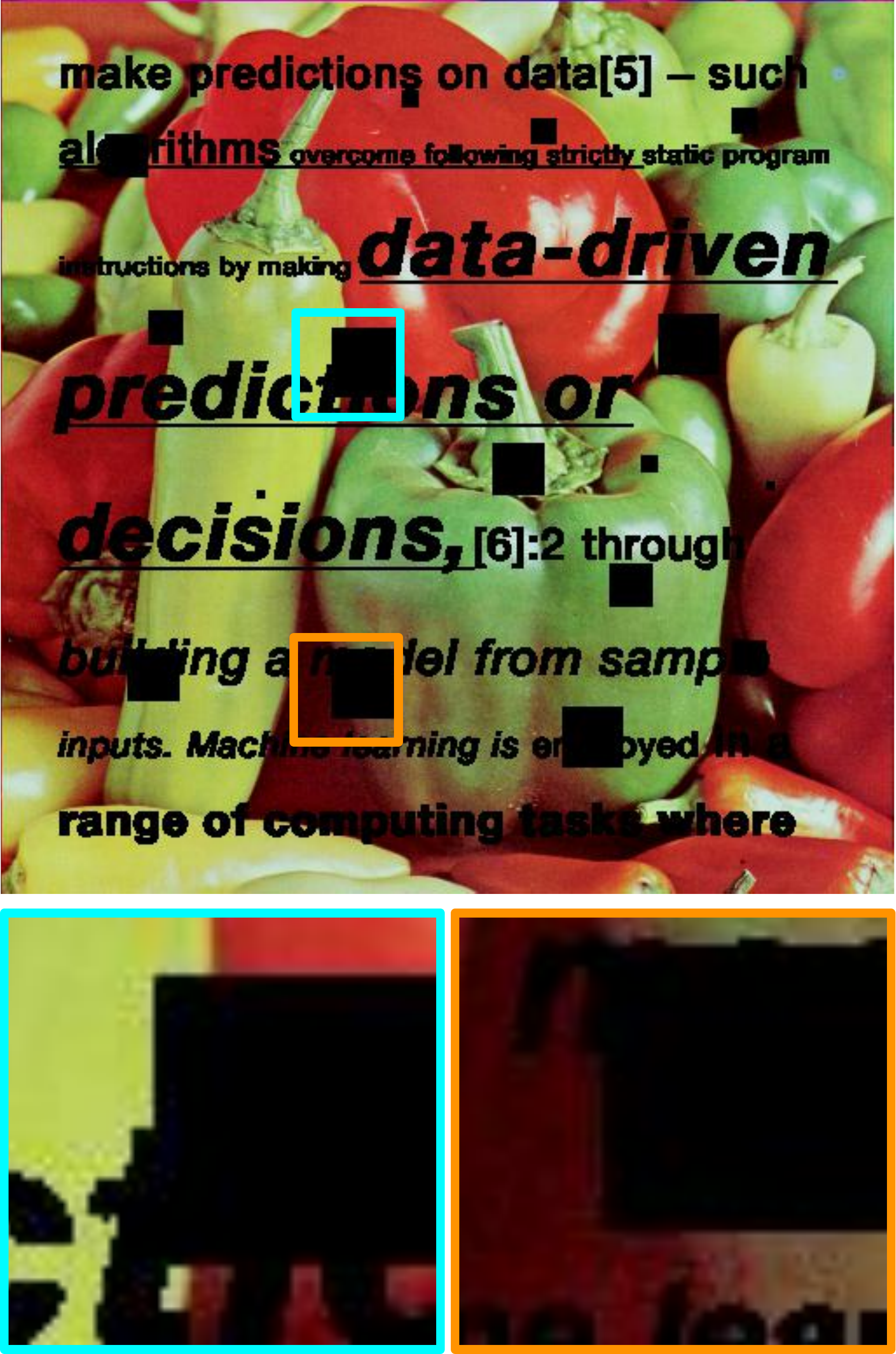} \hspace{-4mm}
			&\includegraphics[width = 0.19\linewidth]{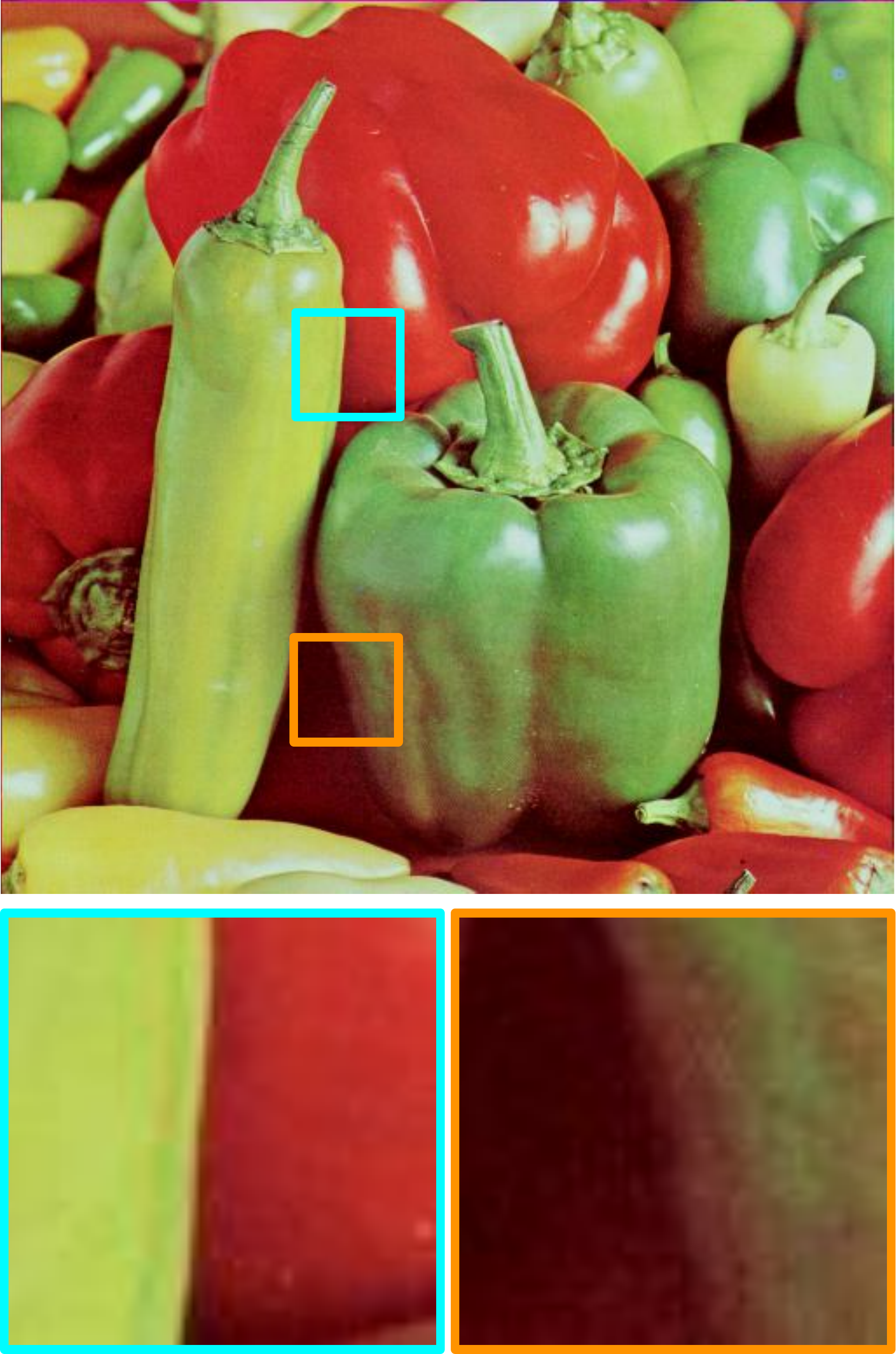} \hspace{-4mm}
			&\includegraphics[width = 0.19\linewidth]{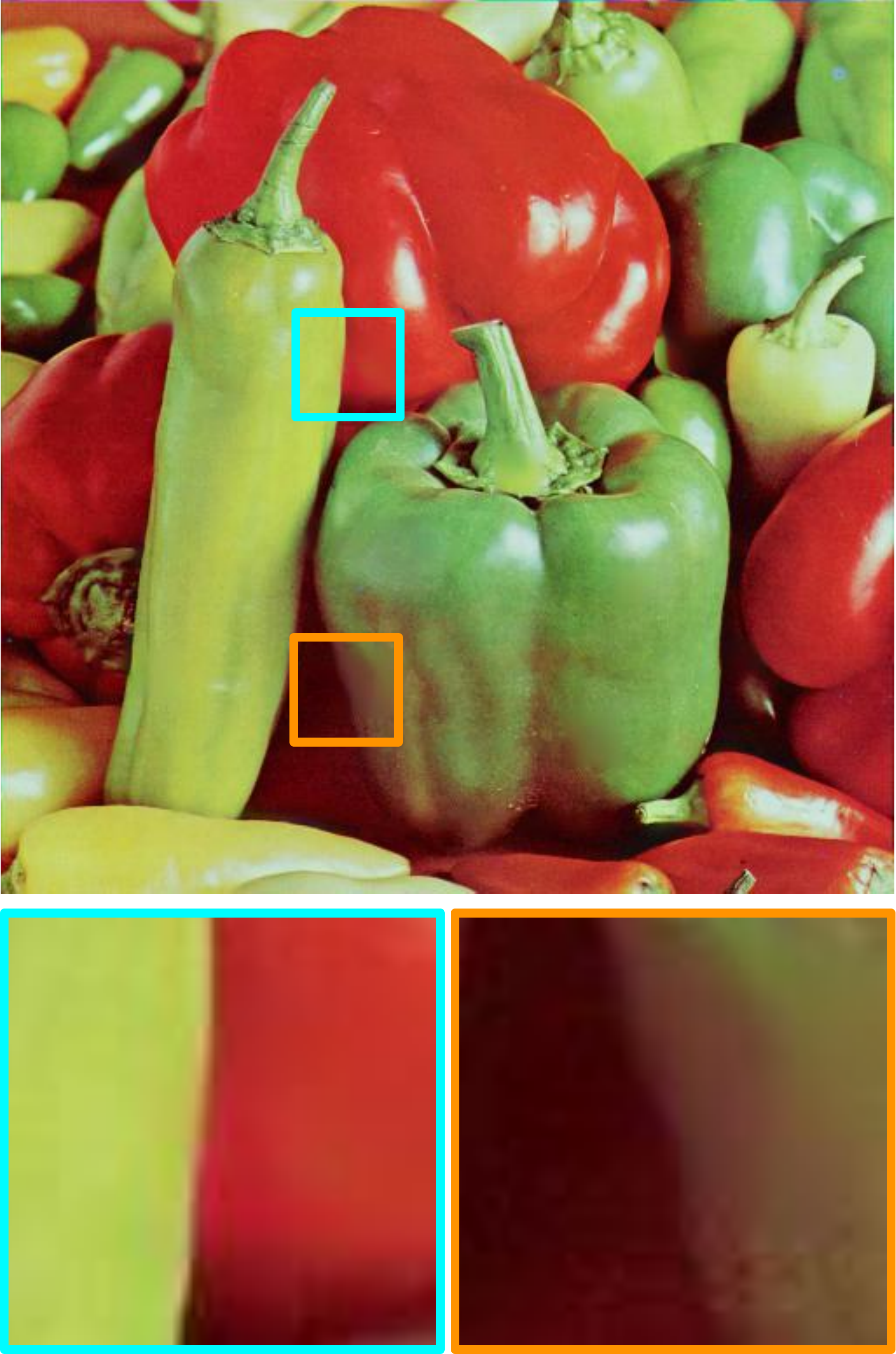} \hspace{-4mm}
			&\includegraphics[width = 0.19\linewidth]{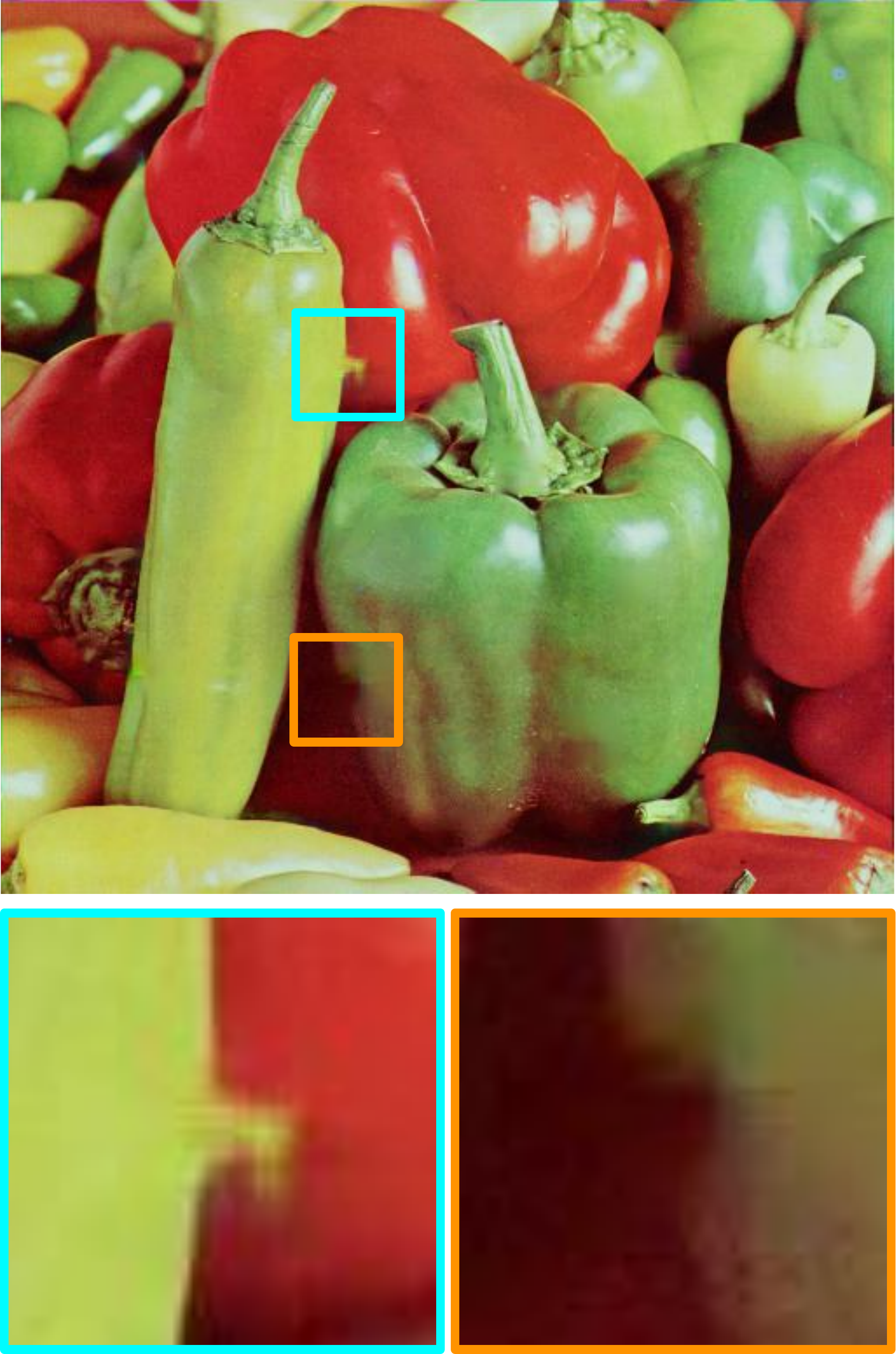} \hspace{-4mm}
			&\includegraphics[width = 0.19\linewidth]{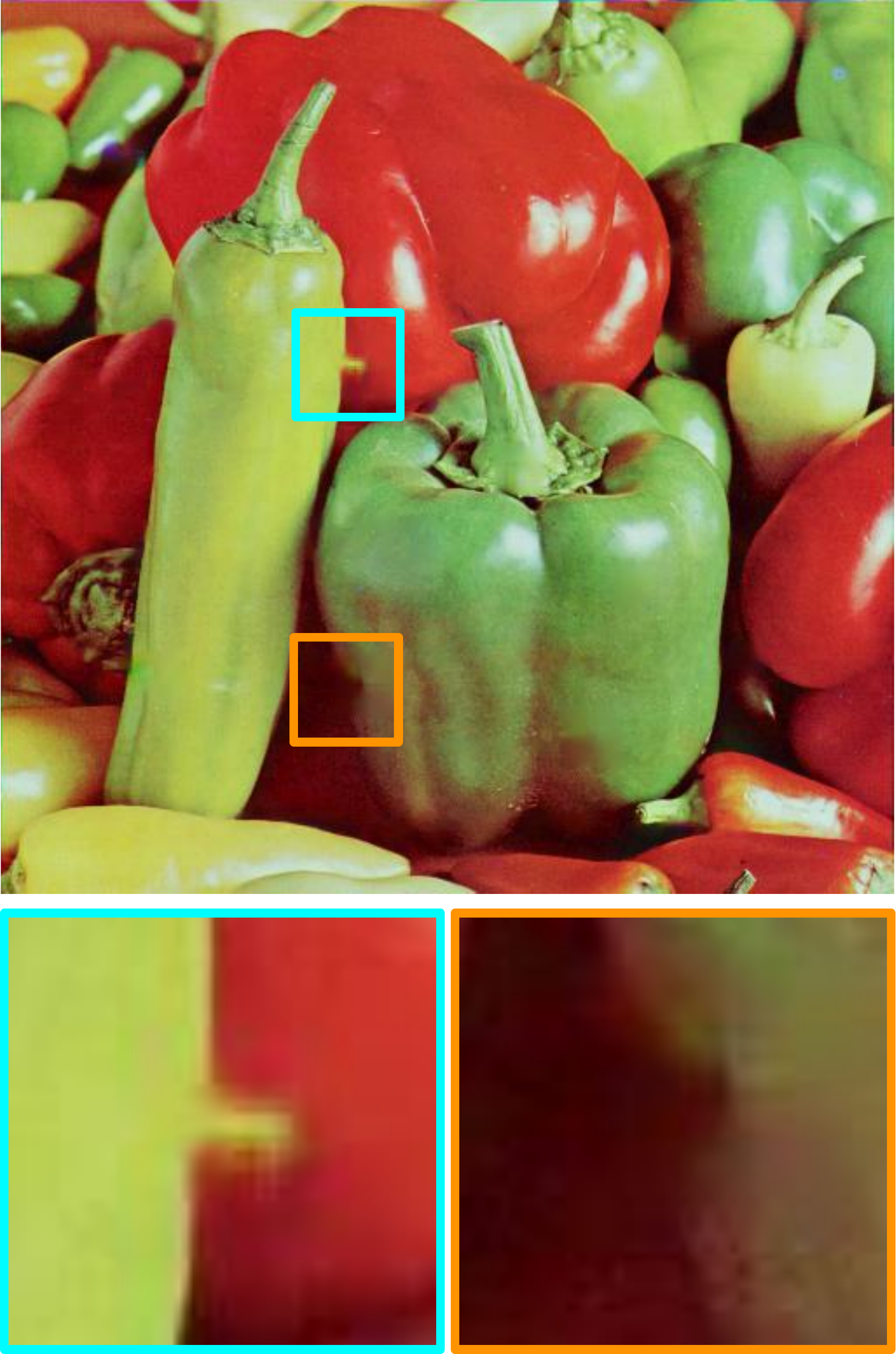}\\
			(PSNR, SSIM) \hspace{-4mm} &(-, -) \hspace{-4mm} &({\bfseries 35.7731}, {\bfseries 0.9615}) \hspace{-4mm} &(34.6972, 0.9585) \hspace{-4mm} &(34.6009, 0.9585)\\
			(a) Input \hspace{-4mm} &(b) Ground Truth \hspace{-4mm} &(c) Net-D+GP+Net-E+$L_1$ \hspace{-4mm} &(d) GP+Net-E+$L_1$ \hspace{-4mm} &(e) Net-E$^{64}$+GP+Net-E+$L_1$\\
		\end{tabular}
	\end{center}
	\caption{Visual comparisons of employing different priors. Net-D can help the estimation of structure and detail.}
	\label{fig: net-d visualization}
\end{figure*}

\subsection{Comparisons with State-of-the-Art Methods}

We present experimental evaluations of the proposed algorithm against several state-of-the-art methods, including blind inpainting methods and even non-blind inpainting methods \cite{Roth:2009iz, Zoran:2011jn, Ren:2015wv}, in terms of their publicly available codes. Table \ref{table: psnr/ssim - Comparisons with  State-of-the-Art Methods} summarizes the quantitative evaluations on several benchmark datasets, and our method performs favorably against previous methods even non-blind methods by a large margin under the measurements PSNR and SSIM.

{\flushleft {\bfseries Running time.}} We benchmark the running time of all methods on dataset Set5 (by grayscale) with a machine of an Intel Core i7-6850K CPU and an NVIDIA GTX 1080 Ti GPU. Table \ref{tab: state-of-the-art run-time} shows that our method runs much faster than other methods. Moreover, it just costs 0.007 seconds to test a grayscale image of size 512 $\times$ 512, e.g., the image baby\_GT.bmp in Set5, making our model capable of inpainting image in real-time.

{\flushleft {\bfseries Visual Comparisons.}} Figure \ref{fig: state-of-the-art visual comparisons} shows several visual comparisons with state-of-the-art methods. Due to the intrinsic difficulty and challenge, previous blind and non-blind methods fail to recover missing information of square regions. Moreover, these methods still contain significant artifacts in dealing with texts. In contrast, our method generates much clearer images. Figure \ref{fig: state-of-the-art visual comparison on an old photograph} shows another set of comparisons tested on an old photograph. This is a real image from \cite{Bertalmio:2000ff}. Our method could generate considerably realistic results.


\section{Analysis and Discussion}
In this section, firstly we further examine the effect of different parts of our proposed network via qualitative and quantitative experiments. 
All of them are conducted in terms of the same parameters described in Section \ref{section: Training Details and Parameters}. 
Then we analyze convergence properties of our networks.
Finally, we discuss the limitations of the proposed method.

For simplicity and clarity, we denote our complete network and method as Net-D+GP+Net-E+$L_1$, where GP is short for pre-extracted gradient prior. 
The network GP+Net-E+$L_1$ denotes training without Net-D, and the network Net-E$^{64}$+GP+Net-E+$L_1$ denotes replacing Net-D with Net-E$^{64}$, where Net-E$^{64}$ is a 20-layer plain network that has the same architectures to those of Net-E except that its last layer has 64 channels. 
Another two networks Net-D+Net-E+$L_1$ and Net-D+GP+Net-E+$L_2$ denotes training without gradient prior and training using $L_2$ loss function, respectively.

\subsection{Effect of Net-D}

We examine the effect of Net-D by replacing it with different sub-networks and compare with GP+Net-E+$L_1$ and Net-E$^{64}$+GP+Net-E+$L_1$.

As mentioned in Section \ref{sec: network architecture}, Net-D is used to capture more useful information and it plays an important role for the inpainting.
Table \ref{table: psnr/ssim of different networks} shows the quantitative comparison between Net-D+GP+Net-E+$L_1$ and GP+Net-E+$L_1$. We can see that the PSNR/SSIM of training with Net-D is significantly higher than that of training without Net-D. A visual comparison is presented in Figure \ref{fig: net-d visualization}. Net-D+GP+Net-E+$L_1$ can recover sharp edges while the results of GP+Net-E+$L_1$ contains some artifacts.


\begin{figure*}[ht]\footnotesize
	\begin{center}
		\begin{tabular}{ccccc}
			\includegraphics[width = 0.19\linewidth]{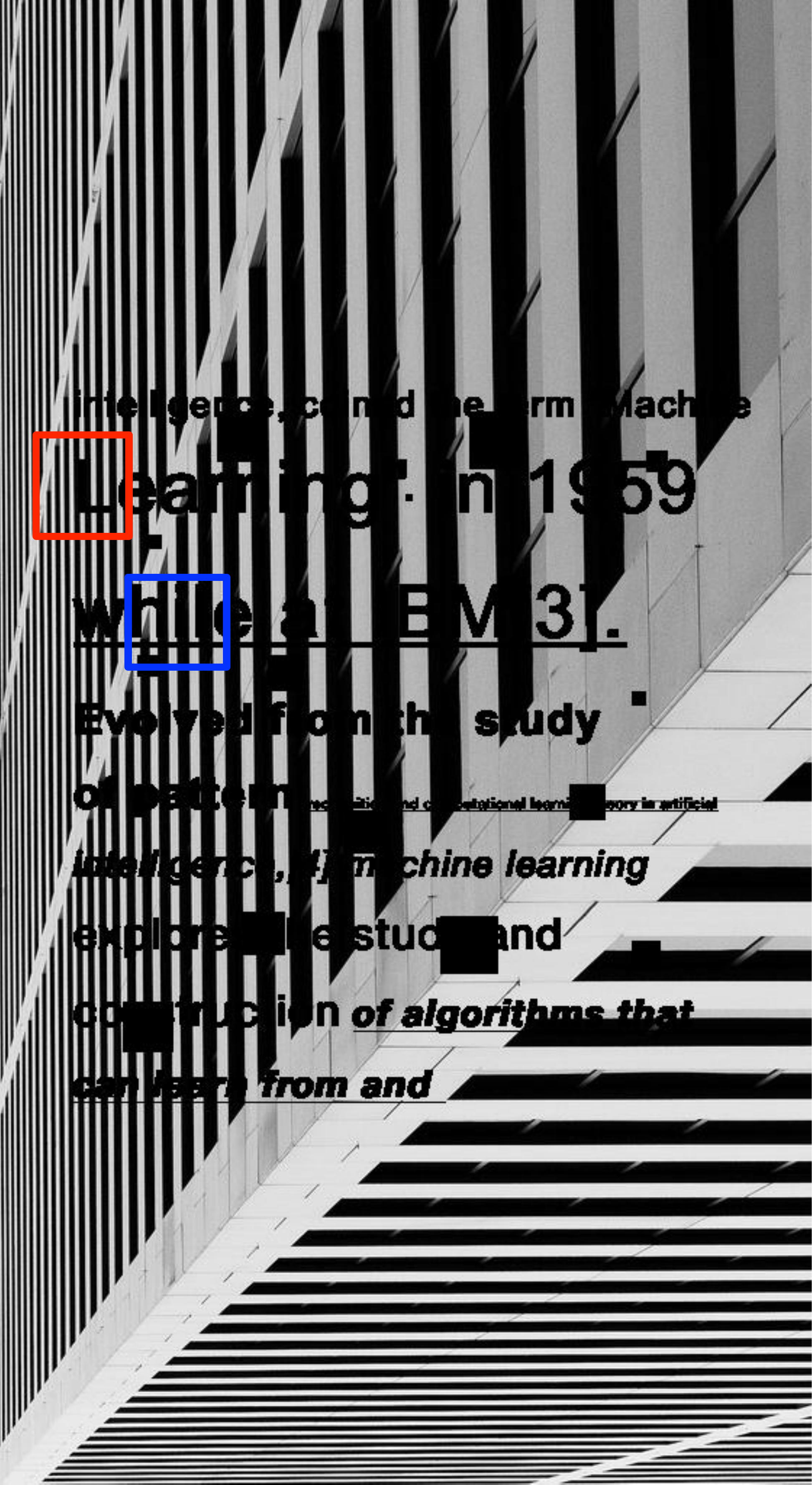} \hspace{-4mm}
			&\includegraphics[width = 0.19\linewidth]{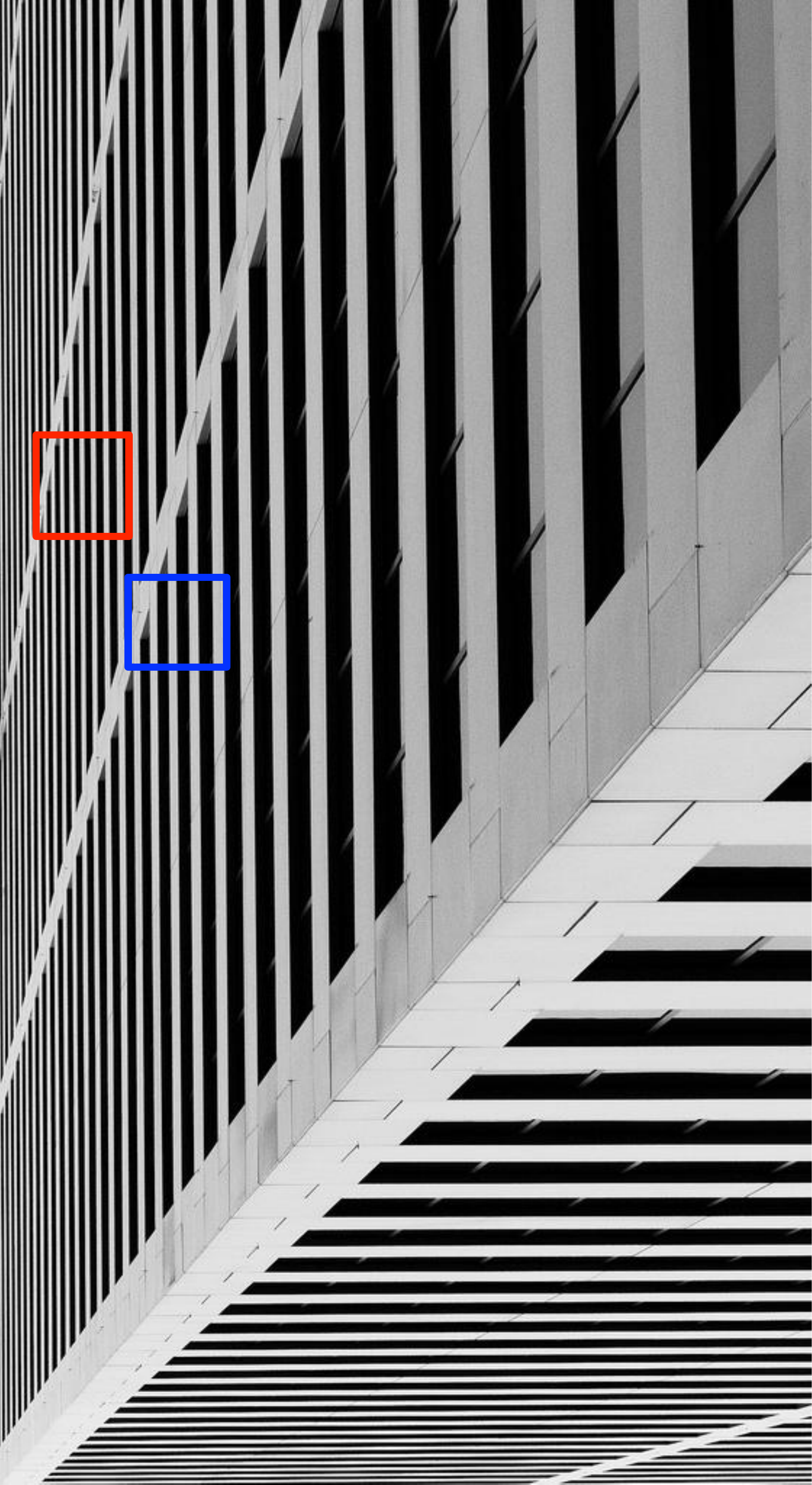} \hspace{-4mm}
			&\includegraphics[width = 0.19\linewidth]{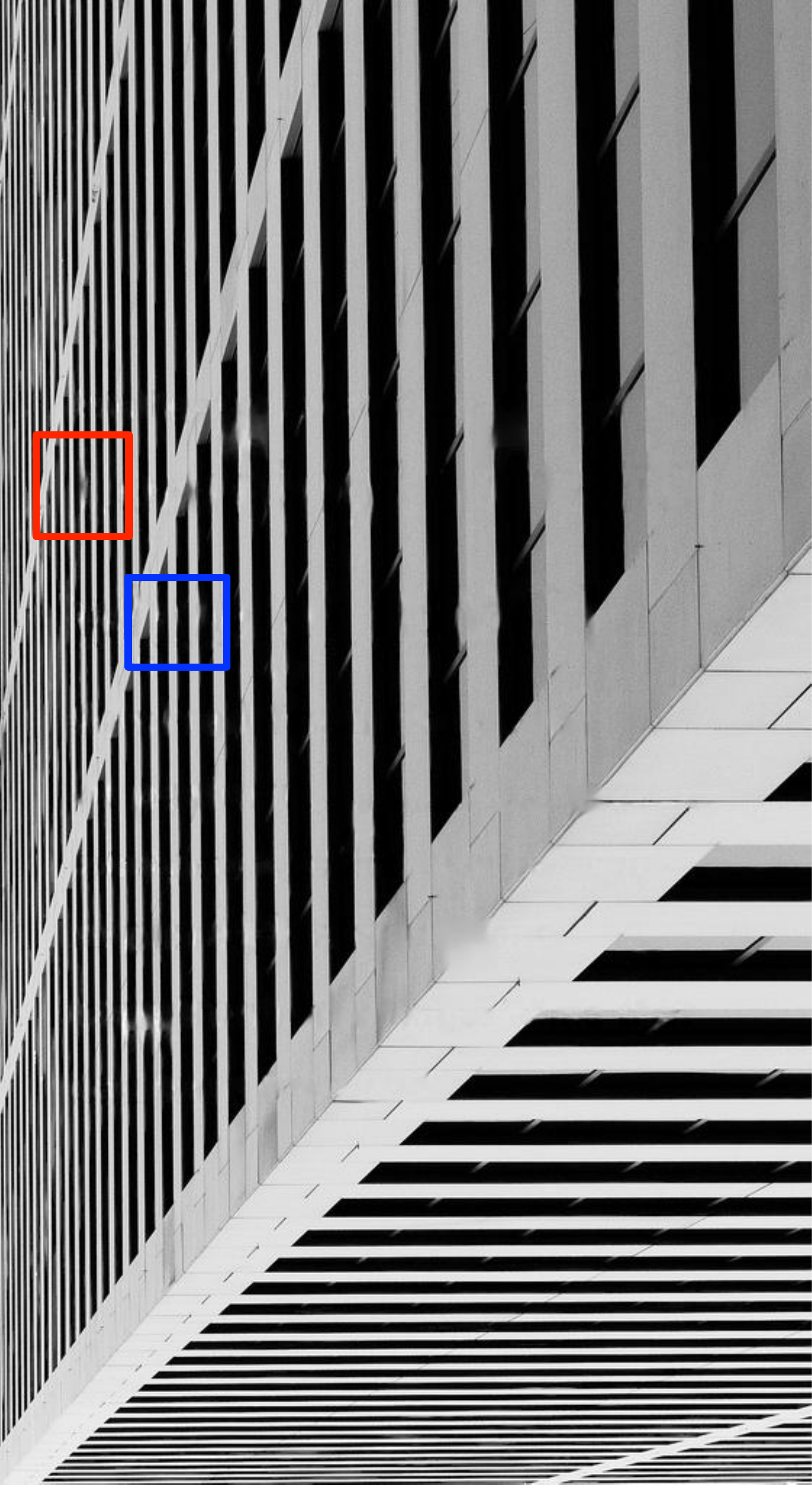} \hspace{-4mm}
			&\includegraphics[width = 0.19\linewidth]{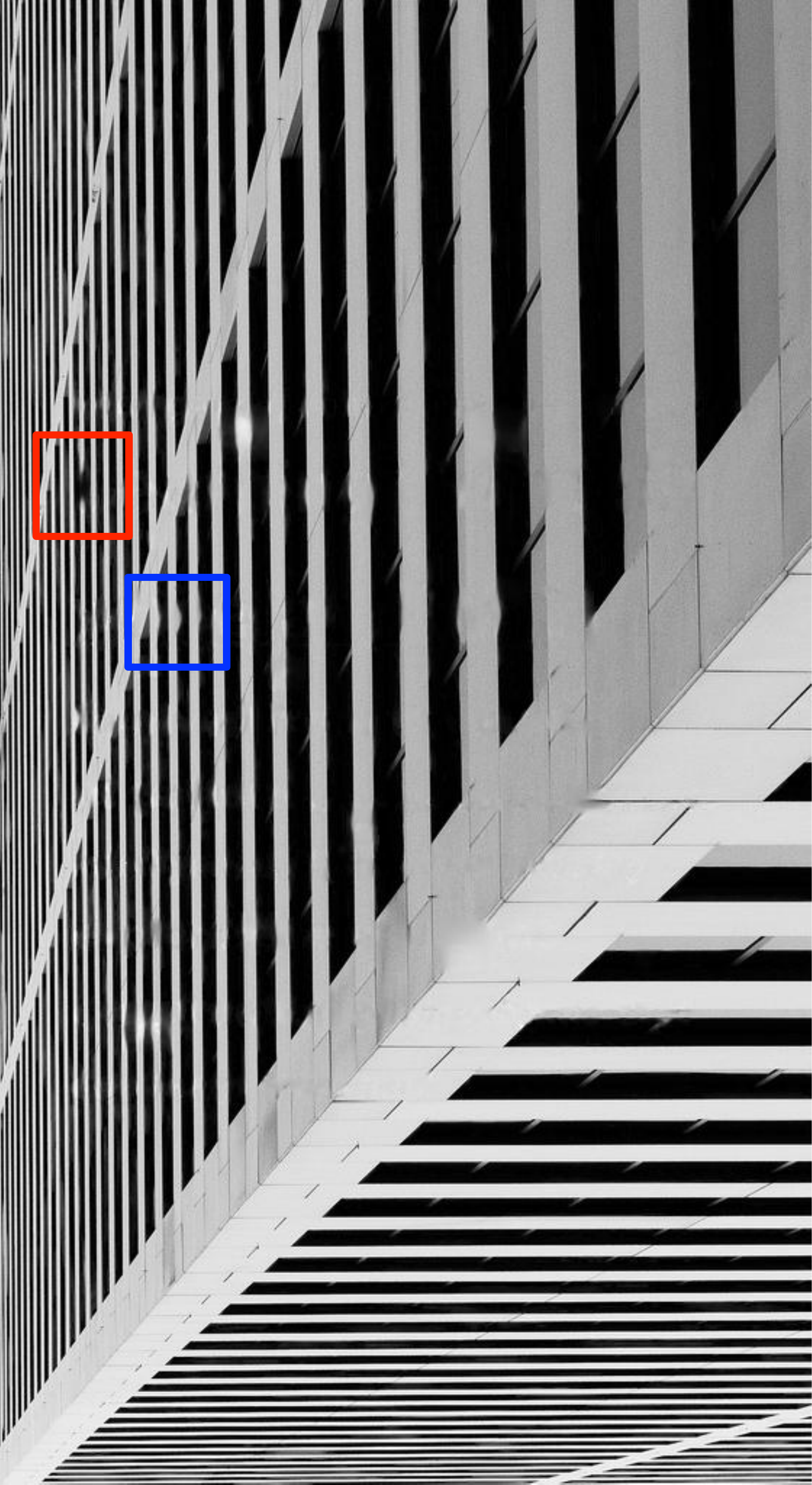} \hspace{-4mm}
			&\includegraphics[width = 0.19\linewidth]{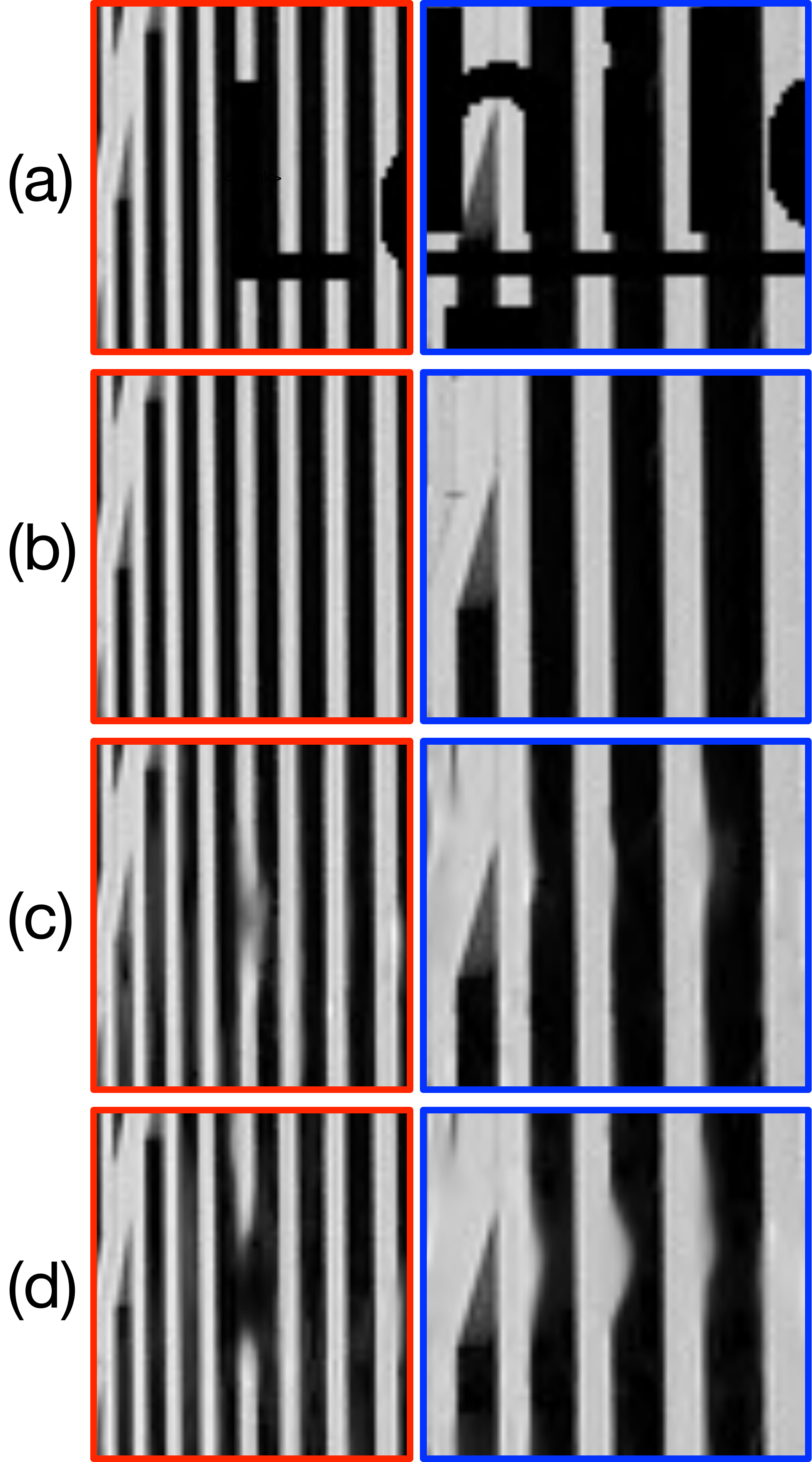} \hspace{-4mm}
			\\
			(PSNR, SSIM) \hspace{-4mm} &(-, -) \hspace{-4mm} &({\bfseries 27.2314}, {\bfseries 0.9752}) \hspace{-4mm} &(27.0494, 0.9618) \hspace{-4mm} &(-, -) \hspace{-4mm}
			\\
			(a) Input \hspace{-4mm} &(b) Ground Truth \hspace{-4mm} &(c) Net-D+GP+Net-E+$L_1$ \hspace{-4mm} &(d) Net-D+Net-E+$L_1$ \hspace{-4mm} &(e) Croped Patches \hspace{-4mm}\\
		\end{tabular}
	\end{center}
	\caption{Visual comparisons of using pre-extracted gradient prior or not.
		The network trained with image gradient is able to recover fine detailed structures while the network trained without gradient fails to fill the useful information in the corrupted regions.}
	\label{figure: gradient prior visualization}
\end{figure*}

\begin{figure}[ht]\footnotesize
	\begin{center}
		\vspace{1mm}
		\hspace{-4.7mm}
		\begin{tabular}{c}
			\vspace{0mm}
			\includegraphics*[width = 0.8\linewidth]{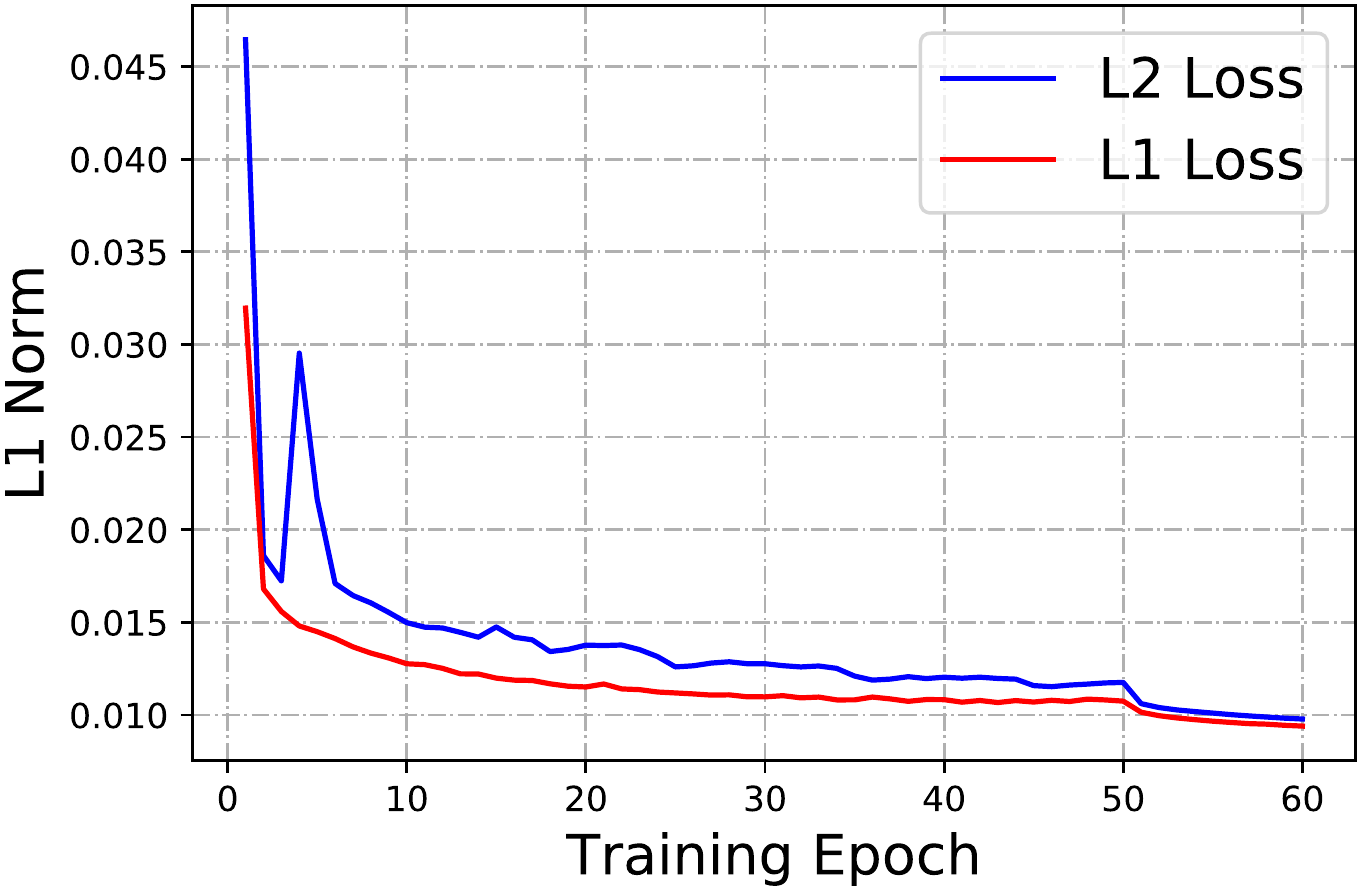}\\
			\hspace{11.4mm} (a) Training Loss \\
			\includegraphics*[width = 0.8\linewidth]{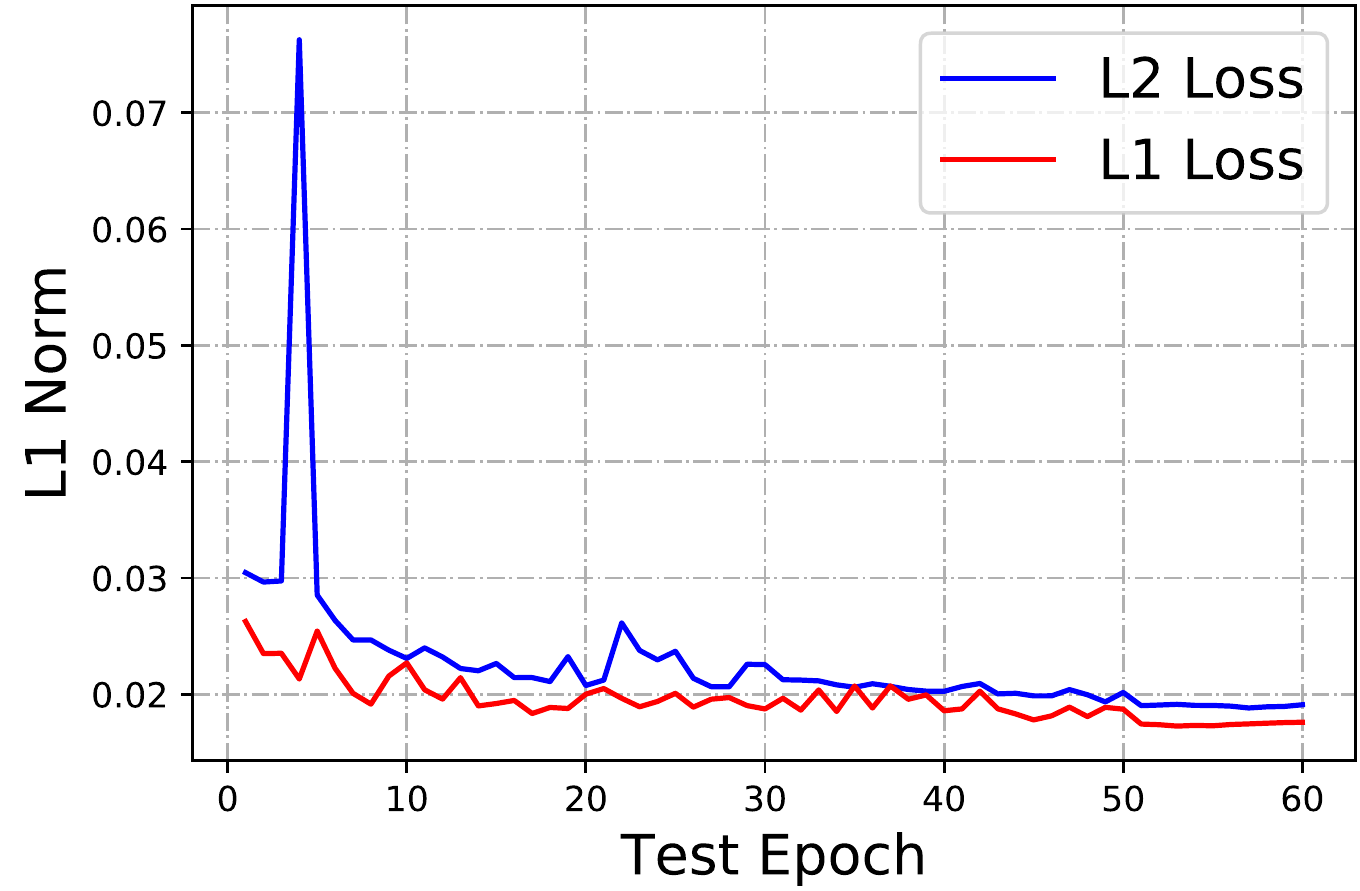}\\
			\hspace{11.4mm} (b) Test Loss \hspace{-20mm}\\
		\end{tabular}
	\end{center}
	\caption{$L_2$/$L_1$ loss in training and test measured by the $L_1$ norm. The error curve of the converged $L_1$ loss function is more stable and lower than that of $L_2$ loss function both in training and test.}
	\label{figure: L2/L1 loss in training and test}
	\vspace{0mm}
\end{figure}

We note that the skip connection operations \cite{Ronneberger:2015vw} play a critical role in Net-D, which are used to copy and concatenate shallow feature maps with deep feature maps. In this way, Net-D could output comprehensive image priors combining low-level features e.g., edges and boundaries, with high-level features e.g., contours and objects.
We compare with Net-E$^{64}$+GP+Net-E+$L_1$, which is a plain network without any skip connection. This network aims to learn high-level semantic features. It is surprising that this network even does not achieve better quantitative evaluations than GP+Net-E+$L_1$. Although Net-E$^{64}$+GP+Net-E+$L_1$ can still remove texts well, it fails to recover sharp edge as better as Net-E+GP+Net-E+$L_1$ when dealing with large squares, suggesting the effectiveness of Net-D.

\subsection{Effect of Pre-extracted Gradient}
\label{ssec: gradient-prior}





The priors learned by Net-D and pre-extracted gradient prior are both employed in our network as showed in Figure \ref{fig: cnn framework}. 
%
%
We use the pre-extract gradients to complement and enhance the features learned by Net-D. 

We train a network without pre-extracted gradient prior named Net-D+Net-E+$L_1$.
Table \ref{table: psnr/ssim of different networks} summarizes the quantitative evaluations of these two networks in terms of PSNR and SSIM, and Net-D+GP+Net-E+$L_1$ achieves higher results than Net-D+Net-E+$L_1$ on the benchmark datasets.
Figure \ref{figure: gradient prior visualization} shows some visual comparisons of using pre-extracted gradient or not. When gradient has been used, the network Net-D+GP+Net-E+$L_1$ is more likely to recover the main structures and generate sharper edges (Figure \ref{figure: gradient prior visualization}(c)), while the network Net-D+Net-E+$L_1$ trained without gradient is less accurate (Figure \ref{figure: gradient prior visualization}(d)), suggesting the effectiveness of pre-extracted gradient.

\begin{figure*}[htbp]\footnotesize
	\begin{center}
		\begin{tabular}{ccccc}
			\includegraphics[width = 0.245\linewidth]{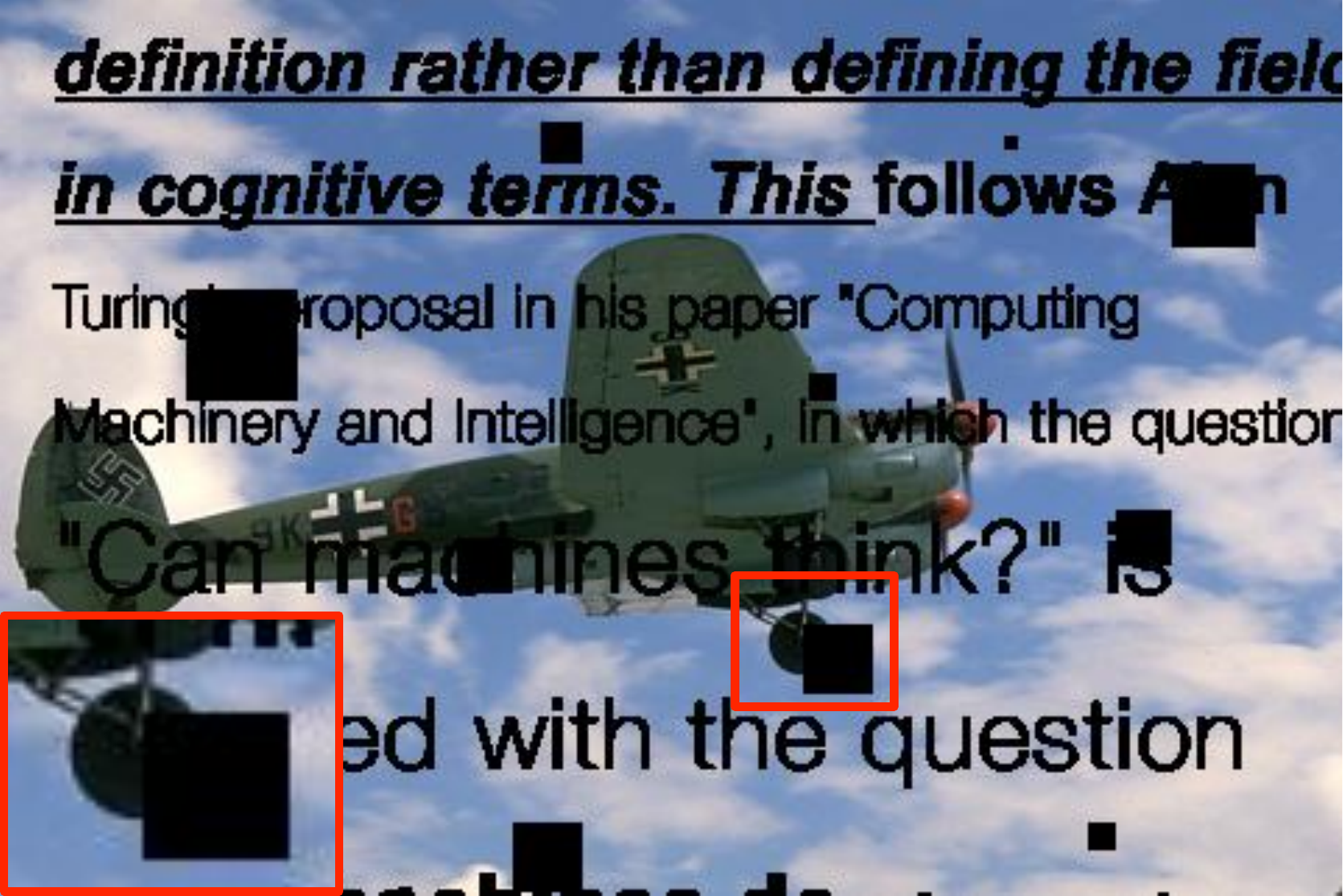} \hspace{-4.5mm}
			&\includegraphics[width = 0.245\linewidth]{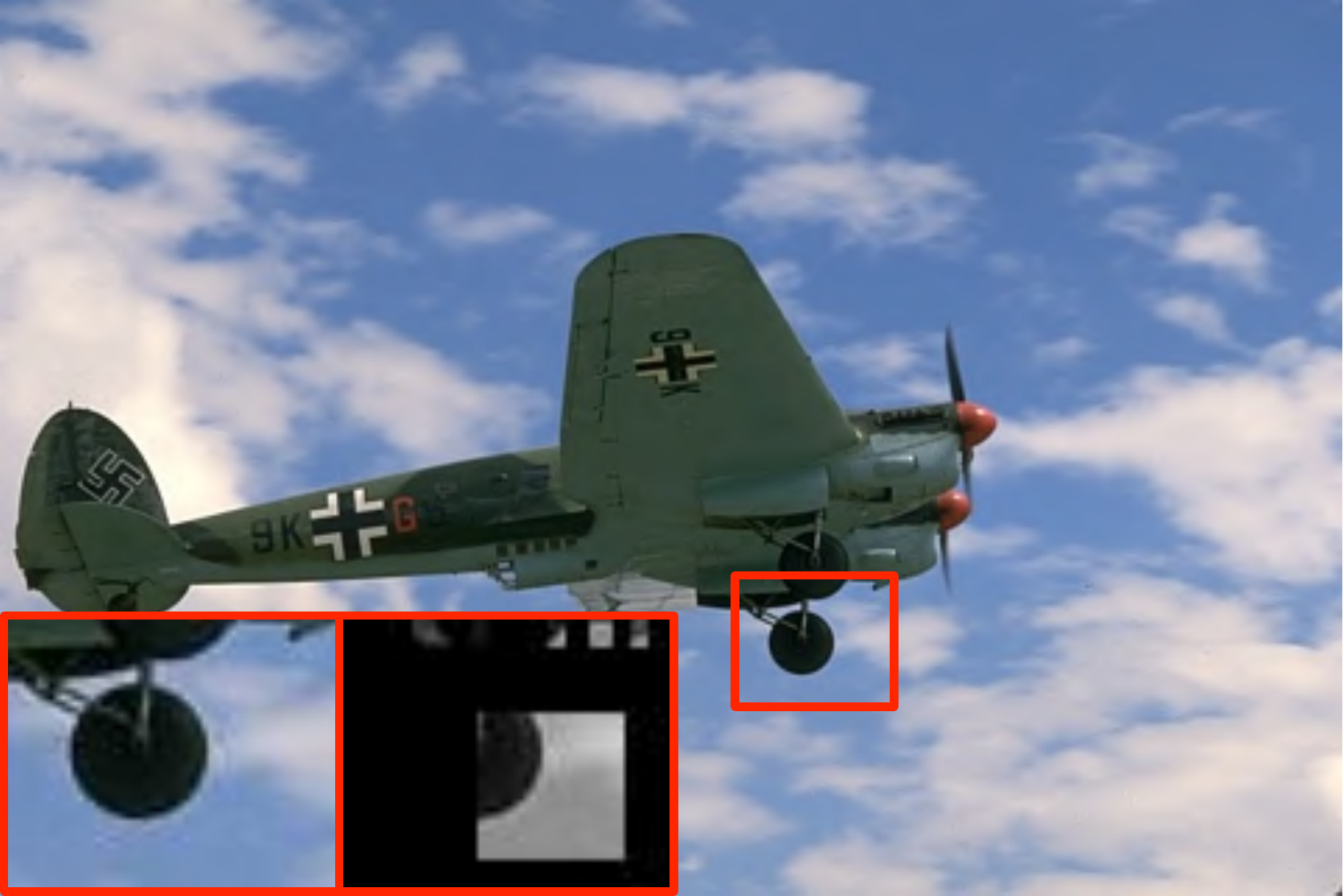} \hspace{-4.5mm}
			&\includegraphics[width = 0.245\linewidth]{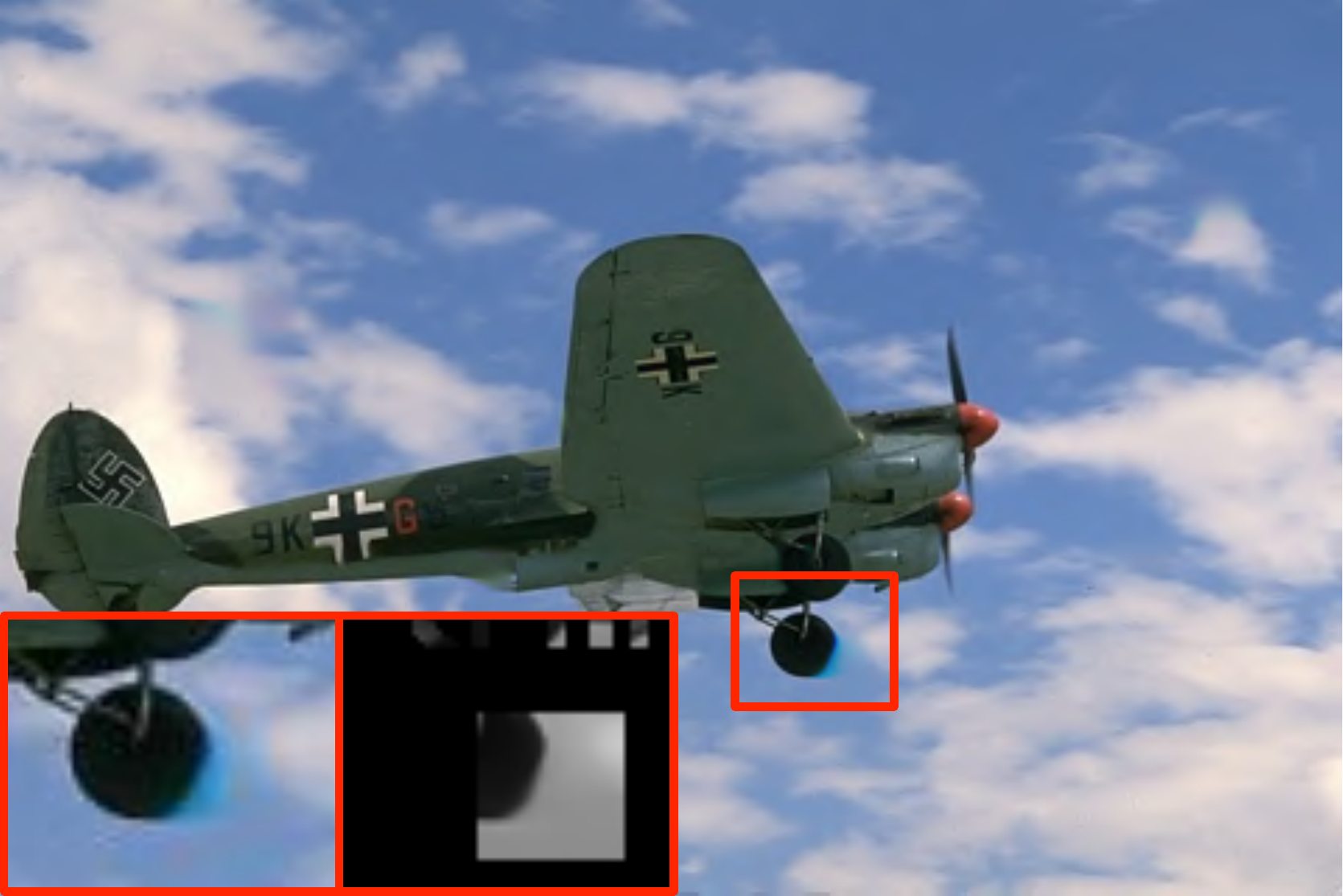} \hspace{-4.5mm}
			&\includegraphics[width = 0.245\linewidth]{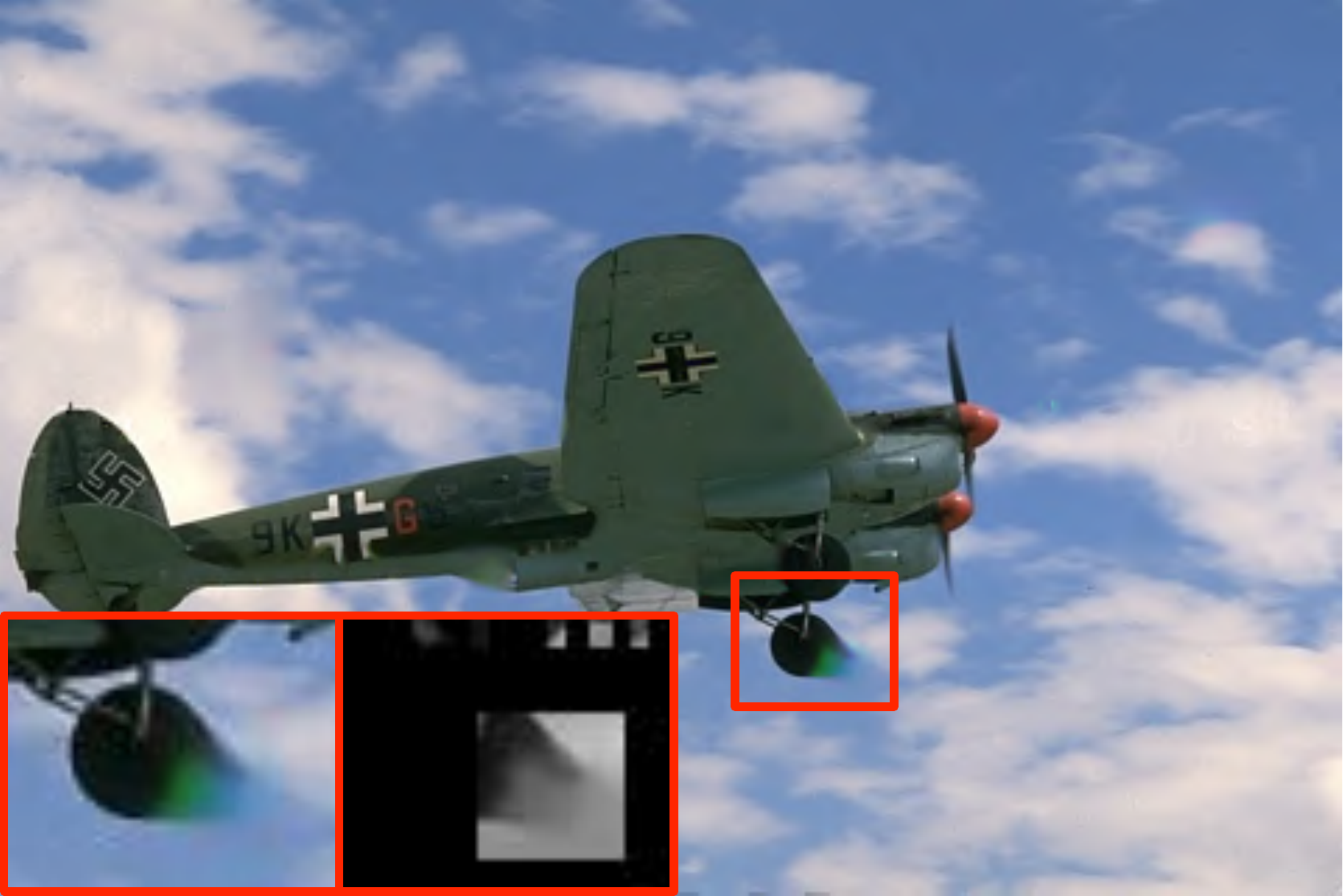} \hspace{-4.5mm}
			&
			\\
			(PSNR, SSIM) \hspace{-4.5mm} &(-, -) \hspace{-4.5mm} &({\bfseries 37.9356}, {\bfseries 0.9845}) \hspace{-4.5mm} &(36.9683, 0.9803) \hspace{-4.5mm}&
			\\
			(a) Input \hspace{-4.5mm} &(b) Ground Truth \hspace{-4.5mm} &(c) Net-D+GP+Net-E+$L_1$ \hspace{-4.5mm} &(d) Net-D+GP+Net-E+$L_2$ \hspace{-4.5mm}&\\
		\end{tabular}
	\end{center}
	\caption{Visual comparisons between $L_1$ and $L_2$ loss function. The second close-ups in Figure \ref{figure: L1-L2 visualization}(b)-(d) denote the ground truth residual image, the residual image learned by $L_1$ norm, and the residual image learned by $L_2$ norm, respectively.}
	\label{figure: L1-L2 visualization}
\end{figure*}

\subsection{Effect of the Loss Function}
\label{ssec: Effect of the Loss Function}
\begin{figure}[ht]\footnotesize
	\begin{center}
		\hspace{-4mm}
		\begin{tabular}{c}
			\includegraphics*[width = 0.99\linewidth]{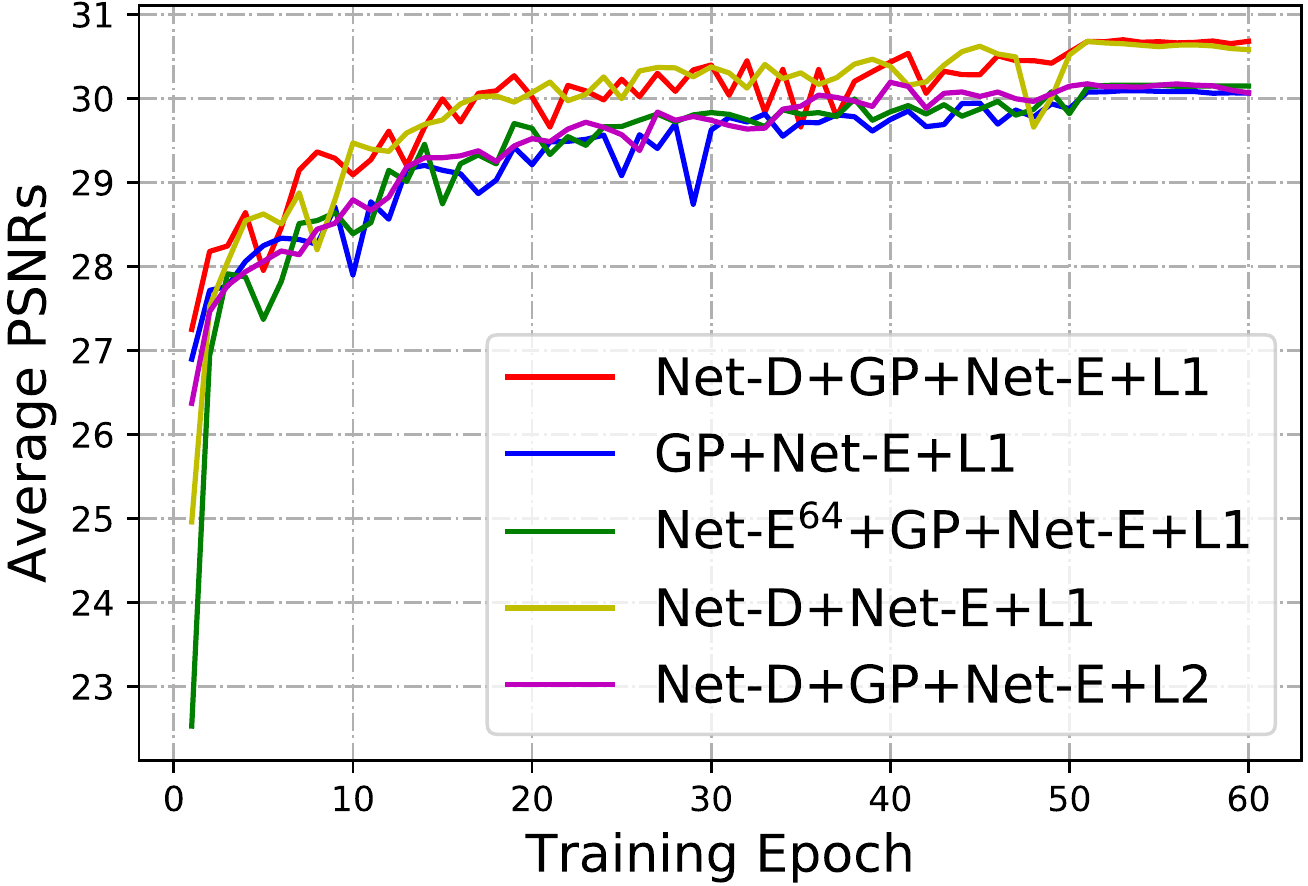}
			\\
		\end{tabular}
	\end{center}
	\caption{Quantitative evaluation of the convergence property on the dataset Set14.}
	\label{figure: Convergence property}
\end{figure}

We note that $L_2$ norm based loss function is widely used in CNN to restore image\cite{Dong:2016fd}.
However, As most information is missing in the corrupted regions, e.g., the tagged region in Figure \ref{figure: L1-L2 visualization}(a), lots of outliers may be introduced. 
Thus the generated image obtained by $L_2$ loss function contains significant artifacts (Figure~\ref{figure: L1-L2 visualization}(d)).

To examine the proposed loss function in image inpainting, we train a network Net-D+GP+Net-E+$L_2$ using the $L_2$ loss function. 
Figure \ref{figure: L1-L2 visualization}(d) shows a recovered image. We can see that the resulting image contains undesirable artifacts and the learned grayscale residual image (the second close-up in Figure \ref{figure: L1-L2 visualization}(d)) is significantly different from the ground truth.

Under these observations, $L_1$ loss function as depicted in (\ref{formula: L1 loss function}) is adopted in this work to alleviate the effect of outliers. As a contrast, the restored image in Figure \ref{figure: L1-L2 visualization}(c) is sharper, and the resulting grayscale residual image gets much closer to the ground truth.
Figure \ref{figure: L2/L1 loss in training and test} shows the values of the training loss functions based on $L_1$ norm and $L_2$ norm. The method with the $L_1$-norm-based reconstruction error converges better than that with the $L_2$-norm-based reconstruction error, both in training and test stage, suggesting that $L_1$ loss function is more robust.
Quantitative evaluations for $L_1$ and $L_2$-norm-based loss function on four benchmark test datasets are summarized in Table \ref{table: psnr/ssim of different networks}.
Net-D+GP+Net-E+$L_1$ has higher PSNR and SSIM values than those of Net-D+GP+Net-E+$L_2$.

\subsection{Convergence Property}
We quantitatively evaluate the convergence properties of our method against the other four networks, i.e., GP+Net-E+$L_1$, Net-E$^{64}$+GP+Net-E+$L_1$, Net-D+Net-E+$L_1$ and Net-D+GP+Net-E+$L_2$, on the dataset Set14. Figure \ref{figure: Convergence property} shows the convergence rate on the dataset Set14.
The proposed algorithms converge well after 50 epochs.

\subsection{Limitations}

\begin{figure}[tp]\footnotesize
	\begin{center}
		\hspace{-4.5mm}
		\begin{tabular}{cc}
			\includegraphics*[width = 0.55\linewidth, height = 0.8\linewidth]{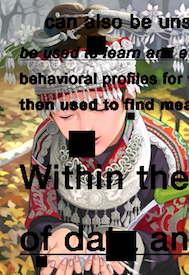} \hspace{-4mm}
			&\includegraphics*[width = 0.55\linewidth, height = 0.8\linewidth]{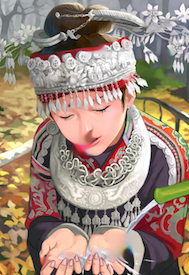} \hspace{-4mm}
			\\
			(a) Input \hspace{-4mm} &(b) Output \hspace{-4mm}
		\end{tabular}
	\end{center}
	\caption{A failure example of our method. The proposed method does not perform well when special structures are corrupted by very large square.}
	\label{figure: limitations}
\end{figure}

The proposed algorithm directly restore a clear image from a corrupted input without any assumptions on the corrupted regions.
However, it does not work well when important structures or details are corrupted. These structures or details are usually unique in an image. 
Figure \ref{figure: limitations} shows a failure example, where the nose and mouse are heavily corrupted, our method fails to recover these two parts. 
Future work will consider using an adversarial learning algorithm to solve this problem.
%


\section{Conclusion}
In this paper, we have proposed an efficient blind image inpainting algorithm that directly restores a clear image from a corrupted input.
We introduce an encoder and decoder architecture to capture useful features and fill semantic information based on the residual learning algorithm. 
We develop $L_1$ loss function in the proposed work and show that it is more robust to outliers.
Extensive qualitative and quantitative experiments testify to the superiority of the proposed method over state-of-the-art algorithms.

{\small
\bibliographystyle{ieee}
\bibliography{egbib}
}

\end{document}